\documentclass[journal]{IEEEtran}
\usepackage{amsfonts}

\usepackage{bm}
\usepackage{times}
\usepackage{epsfig}
\usepackage{graphicx}
\usepackage{amsmath}
\usepackage{amssymb}
\usepackage{subfigure}
\usepackage{multicol}
\usepackage{multirow}
\usepackage{caption}
\usepackage[flushleft]{threeparttable}
\usepackage{color}
\usepackage{multirow}
\usepackage{array}
\usepackage{hyperref}
\usepackage{amssymb}
\usepackage{nccbbb}
\usepackage[ruled,linesnumbered]{algorithm2e}
\usepackage{booktabs}

\begin{document}

\title{Federated Generalized Face Presentation Attack Detection}

\author{Rui Shao, \and
        Pramuditha Perera\textsuperscript{\textsection},  
        Pong C. Yuen,~\IEEEmembership{Senior~Member,~IEEE}, 
        Vishal M. Patel,~\IEEEmembership{Senior~Member,~IEEE}% <-this % stops a space
\thanks{Manuscript received Jun 10, 2021; revised Dec 13, 2021, accepted Apr 25, 2022. This work was partially supported by Research Grants Council (RGC/HKBU12200820), Hong Kong. Vishal M. Patel was supported by the NSF award 1923184. \textit{(Corresponding author: Pong C. Yuen)}}
\thanks{Rui Shao and Pong C. Yuen are with the Department of Computer Science, Hong Kong Baptist University, Hong Kong (e-mail: ruishao@life.hkbu.edu.hk;
pcyuen@comp.hkbu.edu.hk)}
\thanks{Pramuditha Perera is with AWS AI Labs, USA (e-mail: pramudi@amazon.com)}
\thanks{Vishal M. Patel is with Department of Electrical and Computer Engineering, Johns Hopkins University, Baltimore, USA (e-mail: vpatel36@jhu.edu).}}

\markboth{Journal of \LaTeX\ Class Files,~Vol.~14, No.~8, August~2017}%
{Shell \MakeLowercase{\textit{et al.}}: Bare Demo of IEEEtran.cls for IEEE Journals}
% The only time the second header w ill appear is for the odd numbered pages

\maketitle
\begingroup\renewcommand\thefootnote{\textsection}
\footnotetext{This work was conducted prior to joining AWS AI Labs when the author was affiliated with Johns Hopkins University.}
\endgroup

\begin{abstract}
Face presentation attack detection (fPAD) plays a critical role in the modern face recognition pipeline. A fPAD model with good generalization can be obtained when it is trained with face images from different input distributions and different types of spoof attacks. In reality, training data (both real face images and spoof images) are not directly shared between data owners due to legal and privacy issues. In this paper, with the motivation of circumventing this challenge, we propose a Federated Face Presentation Attack Detection (FedPAD) framework that simultaneously takes advantage of rich fPAD information available at different data owners while preserving data privacy. In the proposed framework, each data owner (referred to as \textit{data centers}) locally trains its own fPAD model. A server learns a global fPAD model by iteratively aggregating model updates from all data centers without accessing private data in each of them. Once the learned global model converges, it is used for fPAD inference. To equip the aggregated fPAD model in the server with better generalization ability to unseen attacks from users, following the basic idea of FedPAD, we further propose a Federated Generalized Face Presentation Attack Detection (FedGPAD) framework. A federated domain disentanglement strategy is introduced in FedGPAD, which treats each data center as one domain and decomposes the fPAD model into domain-invariant and domain-specific parts in each data center. Two parts disentangle the domain-invariant and domain-specific features from images in each local data center, respectively. A server learns a global fPAD model by only aggregating domain-invariant parts of the fPAD models from data centers and thus a more generalized fPAD model can be aggregated in server. We introduce the experimental setting to evaluate the proposed FedPAD and FedGPAD frameworks and carry out extensive experiments to provide various insights about federated learning for fPAD.
\end{abstract}

\begin{IEEEkeywords}
	Face presentation attack detection, federated learning, generalization ability
\end{IEEEkeywords}

%%%%%%%%% BODY TEXT
\section{Introduction}

\begin{figure}[!htb]
	
	\begin{center}
		
		\includegraphics[ width=0.9\linewidth]{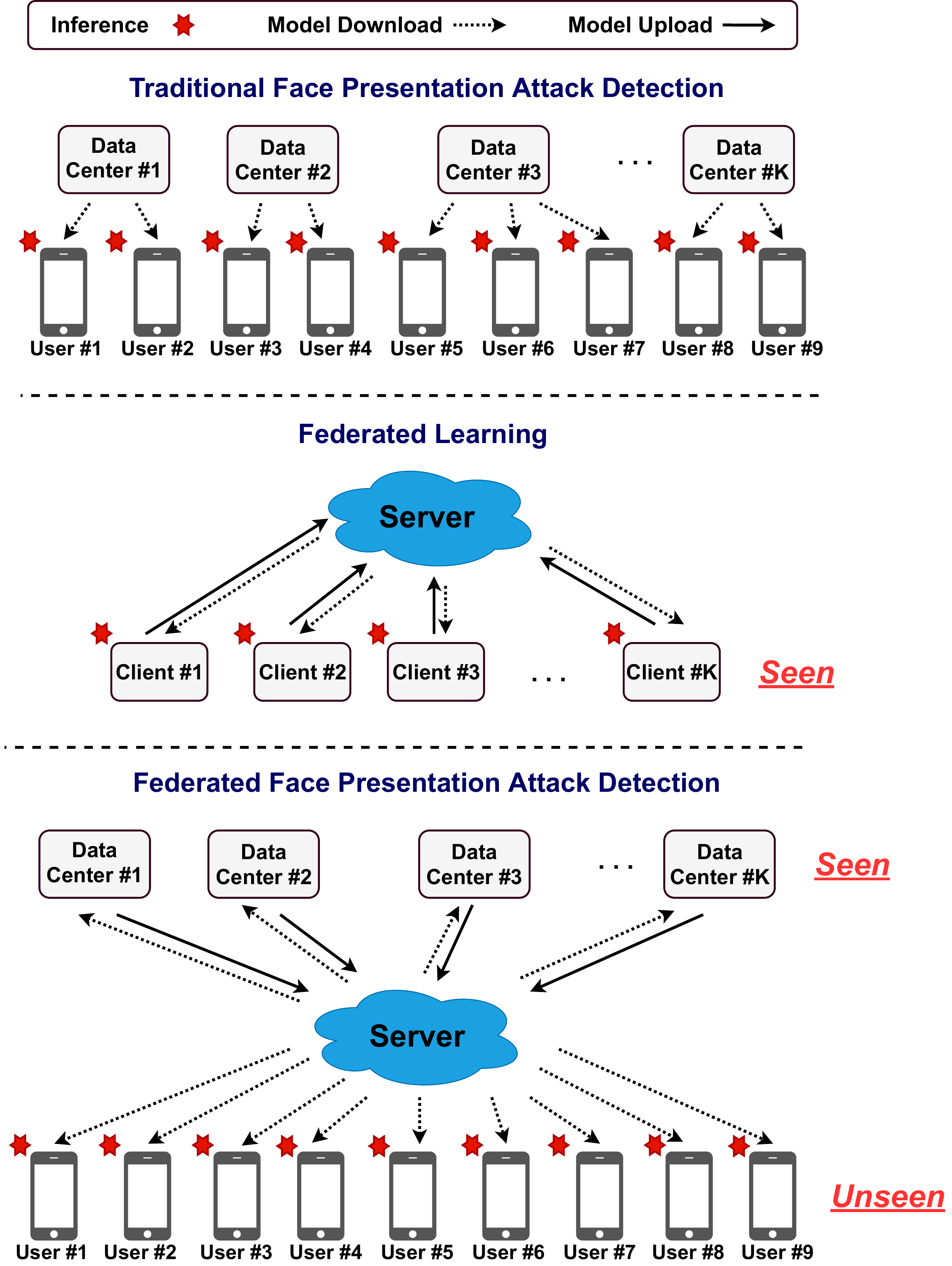}
		
	\end{center}
	
	\caption{Comparison between fPAD (top), traditional FL (middle) and the proposed FedPAD (bottom). FedPAD can be a regarded as a special case of the traditional FL.}
	\label{fig:illustration}
\end{figure}

Recent advances in face recognition methods~\cite{pang2021disp+,pang2021vd} have prompted many real-world applications, such as automated teller machines (ATMs), mobile devices, and entrance guard systems, to deploy this technique as an authentication method. Wide usage of this technology is due to both high accuracy and convenience it provides. However,  many recent works~\cite{2011IJCBmstexture,2016TIFScolortxt,2015TIFSida,RuiShao2018IJCB,2018CVPRauxliary,Shao_2019_CVPR,2018TIFSdynamictext,Shao_2020_AAAI,shao2021federated,shao2020open,shao2022open} have found that this technique is vulnerable to various face presentation attacks such as print attacks, video-replay attacks~\cite{2017FGoulu,2012ICBcasia,2012BIOSIGidiap,2015TIFSida,2018CVPRauxliary} and 3D mask attacks~\cite{2018ECCVrPPG,2016ECCVrPPG,2018TIFSdynamictext,RuiShao2018IJCB}. Therefore, developing face presentation attack detection (fPAD) methods that make current face recognition systems robust to face presentation attacks has become a topic of interest in the biometrics community.

In this paper, we consider the deployment of a fPAD system in a real-world scenario. We identify two types of stakeholders in this scenario -- \textit{data centers} and \textit{users}. \textit{Data centers} are entities that design and collect fPAD datasets and propose fPAD solutions. Typically \textit{data centers} include research institutions and companies that carry out the research and development of fPAD. These entities have access to both \textit{real data} and \textit{spoof data} and therefore are able to train fPAD models. Different \textit{data centers} may contain images of different identities and different types of \textit{spoof data}. However, each \textit{data center} has  limited data availability. Real face images are obtained from a small set of identities and spoof attacks are likely to be from  a few known types of attacks. Therefore, these fPAD models have poor generalization ability~\cite{Shao_2019_CVPR,Shao_2020_AAAI} and are likely to be vulnerable to attacks unseen during training.

On the other hand, \textit{users} are individuals or entities that make use of fPAD solutions. For example, when a fPAD algorithm is introduced in mobile devices, mobile device customers are identified as \textit{users} of the fPAD system.  \textit{Users} have access only to \textit{real data} collected from local devices. Due to the absence of \textit{spoof data}, they cannot locally train fPAD models. Therefore, each \textit{user} relies on a model developed by a \textit{data center} for fPAD as shown in Figure~\ref{fig:illustration} (top). Since \textit{data center} models lack generalization ability, inferencing with these models is likely to result in erroneous predictions. 

It has been shown that utilizing \textit{real data} from different input distributions and \textit{spoof data} from different types of spoof attacks through domain generalization~\cite{2018MMhada,Shao_2019_IETIP} and meta-learning techniques can significantly improve the generalization ability of fPAD models~\cite{Shao_2019_CVPR,Shao_2020_AAAI}. Therefore, the performance of fPAD models, shown in Figure~\ref{fig:illustration} (top), can be improved if data from all \textit{data centers} can be exploited collaboratively. In reality, due to data sharing agreements and privacy policies, \textit{data centers} are not allowed to share collected fPAD data with each other. For example, when a \textit{data center} collects face images from individuals using a social media platform, it is agreed not to share the collected data with third parties. 

In this paper, we present a framework called Federated Face Presentation Attack Detection (FedPAD) based on the principles of Federated Learning (FL) targeting fPAD. The proposed method exploits information across all \textit{data centers} while preserving data privacy. In the proposed framework, models trained at \textit{data centers} are shared and aggregated while training images are kept private in their respective \textit{data centers}, thereby preserving privacy.  

Federate learning is a distributed and privacy preserving machine learning technique~\cite{mcmahan2016communication,li2019federated,smith2017federated,sahu2018convergence,mohri2019agnostic}. FL training paradigm defines two types of roles named \textit{server} and \textit{client}.  \textit{Clients} contain training data and the capacity to train a model. As shown in Fig.~\ref{fig:illustration} (middle), each client trains its own model locally and uploads them to the \textit{server} at the end of each training iteration. \textit{Server} aggregates local updates and produces a global model. This global model is then shared with all clients which will be used in their subsequent training iteration. This process is continued until the global model is converged. During the training process, data of each client is kept private. Collaborative FL training allows the global model to exploit rich local clients information while preserving data privacy.   

In the context of FedPAD, both \textit{data centers} and \textit{users} can be identified as clients. However, roles of \textit{data centers}  and \textit{users} are different from conventional clients found in FL. As illustrated in Fig.~\ref{fig:illustration} (middle), in FL, all \textit{clients} train models and carry out inference locally and clients in the testing are usually seen during the training. Therefore, most current FL work focus on exploiting the \textit{\textbf{personalization}} issue which aims to adapt the aggregated FL model to the seen clients in the testing. In contrast, in FedPAD, only \textit{data centers} carry out local model training. \textit{Data centers} share their models with the \textit{server} and download the global model during training. On the other hand, \textit{users} download the global model at the end of the training procedure and only carry out inference as shown in Figure~\ref{fig:illustration} (bottom). Therefore, the downloaded FL model will inevitably encounter various unseen face presentation attacks from the users. As such, the proposed FedPAD focuses on exploring the \textit{\textbf{generalization}} of FL model which aims to generalize well to unseen attacks from users in the testing. 

To equip the aggregated fPAD model in the server with better generalization ability to unseen attacks from users in the testing, based on the basic concept of FedPAD, a Federated Generalized Face Presentation Attack Detection (FedGPAD) framework is further proposed. Inspired by the current domain generalization methods which improve the generalization ability of models by learning domain-invariant features, we introduce a federated domain disentanglement strategy in FedGPAD. This strategy treats each data center as one source domain and decomposes the fPAD model in every data center into domain-invariant and domain-specific parts. Domain-invariant and domain-specific features can be disentangled from images by these two parts correspondingly. Considering the privacy issues, we carry out domain-invariant feature learning under the FL framework by conducting the communications of model parameters between domain-invariant parts of the fPAD models from the data centers and the server. Therefore, a global fPAD model with more domain-invariant information will be aggregated in the server in a privacy preserving manner, which is able to generalize well to unseen attacks.

Main contributions of this paper can be summarized as follows:

\noindent 1. This paper is the first to study the FL technique for the task of fPAD. Compared to the most existing FL works exploiting the \textit{\textbf{personalization}} issue, this paper studies the \textit{\textbf{generalization}} issue and thus proposes a Federated Face Presentation Attack Detection (FedPAD) framework to develop a generalized fPAD model in a data privacy preserving way. 

\noindent 2. To equip the fPAD model with better generalization ability to unseen attacks from users under FL framework, we further propose a Federated Generalized Face Presentation Attack Detection (FedGPAD) framework. A federated domain disentanglement strategy is introduced in this framework, contributing to a global fPAD model with more domain-invariant information in the server.

\noindent 3. An experimental setting is defined for the FL framework for the task of fPAD.  Extensive experiments are carried out to show the effectiveness of the proposed framework. Various issues and insights about FL for fPAD are also discussed.

\section{Related Work}

\subsection{Face Presentation Attack Detection}
Current fPAD methods can be categorized under single-domain and multi-domain approaches. The single-domain approach focuses on extracting discriminative cues between real and spoof samples from a single dataset, which can be further divided into appearance-based methods and temporal-based methods. Appearance-based methods focus on extracting various discriminative appearance cues for detecting face presentation attacks. Multi-scale LBP~\cite{2011IJCBmstexture} and color textures~\cite{2016TIFScolortxt} methods are two typical texture-based methods that extract various LBP descriptors in various color spaces for the differentiation between real/spoof samples. Image distortion analysis~\cite{2015TIFSida} aims to detect the surface distortions as the discriminative cue. Moreover, Yu \emph{et al.}~\cite{yu2020searching} proposes Central Difference Convolution (CDC) to capture intrinsic detailed patterns via aggregating both intensity and gradient information. Then a Bilateral Convolutional Networks~\cite{yu2020face} is proposed to capture intrinsic material-based patterns via aggregating multi-level bilateral macro- and micro- information. On the other hand, temporal-based methods extract different discriminative temporal cues through multiple frames between real and spoof samples. Various dynamic textures are exploited in~\cite{2014EJIVPlbptop,2018TIFSdynamictext,RuiShao2018IJCB} to extract discriminative facial motions. rPPG signals are exploited by Liu \emph{et al.}~\cite{2016ECCVrPPG,2018ECCVrPPG} to capture discriminative heartbeat information from real and spoof videos. ~\cite{2018CVPRauxliary} learns a CNN-RNN model to estimate different face depth and rPPG signals between the real and spoof samples. Spatio-temporal information of fPAD is also exploited in~\cite{wang2020deep} by using Spatio-Temporal Propagation Module (STPM). Lazaro \emph{et al.}~\cite{gonzalez2020fisher} explore a feature space based on Fisher vectors. Compact Binarised Statistical Image Features (BSIF) histograms are calculated to search this feature space. In such a feature space, semantic feature subsets from known samples can be found to enhance the detection of unknown attacks. The method proposed in~\cite{thummel2021facial} exploits a sequence of 3D face scans to capture the plausibility of facial behavior. Temporal curvature change is further used to measure facial behavior with a compact feature representation.

Various fPAD datasets are introduced recently that explore different characteristics and scenarios of face presentation attacks. Treating each dataset as one domain, the multi-domain approach is proposed to improve the generalization ability of the fPAD model to unseen attacks. Recent work~\cite{Shao_2019_CVPR} casts fPAD as a domain generalization problem and proposes a multi-adversarial discriminative deep domain generalization framework to search generalized differentiation cues in a shared and discriminative feature space among multiple fPAD datasets. ~\cite{liu2019deep} treats fPAD as a zero-shot problem and proposes a Deep Tree Network to partition the spoof samples into multiple sub-groups of attacks. ~\cite{Shao_2020_AAAI,qin2020learning} addresses fPAD based on a meta-learning framework and enables the model to learn to generalize well through simulated train-test splits among multiple datasets. Moreover, Wang \emph{et al.}~\cite{wang2020cross} proposes a disentangled representation learning (DR-Net) and a multi-domain learning (MD-Net) to learn the disentangled representation learning for cross-domain presentation attack detection. Lately, Yu \emph{et al.}~\cite{yu2020fas} exploit the neural architecture search (NAS) for fPAD and propose a cross domain/type-aware meta-learning to form a task-aware network. These multi-domain methods have access to data from multiple datasets or multiple spoof sub-groups that enable them to obtain generalized models. In this paper, we study the scenario in which each \textit{data center} contains data from a single domain. Due to data privacy issues, we assume that they do not have access to data from other \textit{data centers} and thus aim to exploit multi-domain information in a privacy preserving manner.

\subsection{Federated Learning}
Federated learning is a decentralized machine learning approach that enables multiple local clients to collaboratively learn a global model with the help of a server while preserving data privacy of local clients. Federated averaging (FedAvg)~\cite{mcmahan2016communication}, one of the fundamental frameworks for FL, learns a global model by averaging model parameters from local clients. FedProx~\cite{sahu2018convergence} and Agnostic Federated Learning (AFL)~\cite{mohri2019agnostic} are two variants of FedAvg which aim to address the bias issue of the learned global model towards different clients. These two methods achieve better models by adding a proximal term to the cost functions and optimizing a centralized distribution mixed with client distributions, respectively. More recently, Federated Matched Averaging (FedMA)~\cite{wang2020federated} algorithm is proposed to improve FedAvg with a coordinate-wise averaging of weights, which exploits permutation invariance of the neurons of a network. In addition, several personalized variants of FL have been proposed to exploit the issue of  \textit{\textbf{personalization}}  of FL~\cite{dinh2020personalized,fallah2020personalized}, which aim to adapt the FL model to each local client during testing. However, these local clients during testing are all seen during training. In a practical scenario of fPAD, many unseen face presentation attacks will be presented to users during testing. Therefore, this paper attempts to address the issue of \textit{\textbf{generalization}} of FL, with the objective of improving the generalization ability of fPAD under the FL framework.

\subsection{Disentanglement Learning}
The objective of disentanglement learning is to factorize the data into distinct informative factors of variations~\cite{locatello2019challenging} so that complex data can be better represented by the disentangled features. In facial recognition, DR-GAN~\cite{tran2017disentangled} disentangles a facial image into identity and pose factors which can be used for pose-invariant face recognition. The method proposed in~\cite{xiao2018elegant} disentangles a facial image into different factors to encodes distinct facial attributes respectively. Moreover, ~\cite{zhang2019gait} disentangles an input gait video into different representations of appearance, canonical, and pose features for gait recognition. Image synthesis method ~\cite{esser2018variational} utilizes U-Net and Variational Auto Encoder (VAE) to disentangles an image into appearance and shape.

Recently, some works have also exploited the disentanglement learning for fPAD~\cite{zhang2020face,liu2020disentangling}. In particular, \cite{liu2020disentangling} designs an adversarial learning framework to disentangle the spoof traces from input faces for fPAD and further synthesizes realistic new spoof faces. ~\cite{zhang2020face} proposes to disentangle the liveness features and content features from facial images, and the liveness features are further used for fPAD. There are some key differences between the above two fPAD methods and the proposed FedGPAD framework in this paper. First, compared with disentangling spoof traces~\cite{liu2020disentangling} or liveness features~\cite{zhang2020face} within a single source domain, FedGPAD focuses on disentangling the domain-invariant and domain-specific features with the objective of improving the generalization ability to unseen face presentation attacks with multi-source domain (data centers) data. Second, the above two fPAD methods carry out disentanglement learning without considering the privacy issue. In contrast, the proposed federated domain disentanglement in FedGPAD is carried out under the FL framework, which is the first to conduct the domain-invariant feature learning in a privacy preserving way.

\section{Proposed Method}

\begin{figure}[!htb]
	
	\begin{center}
		
		\includegraphics[width=1\linewidth]{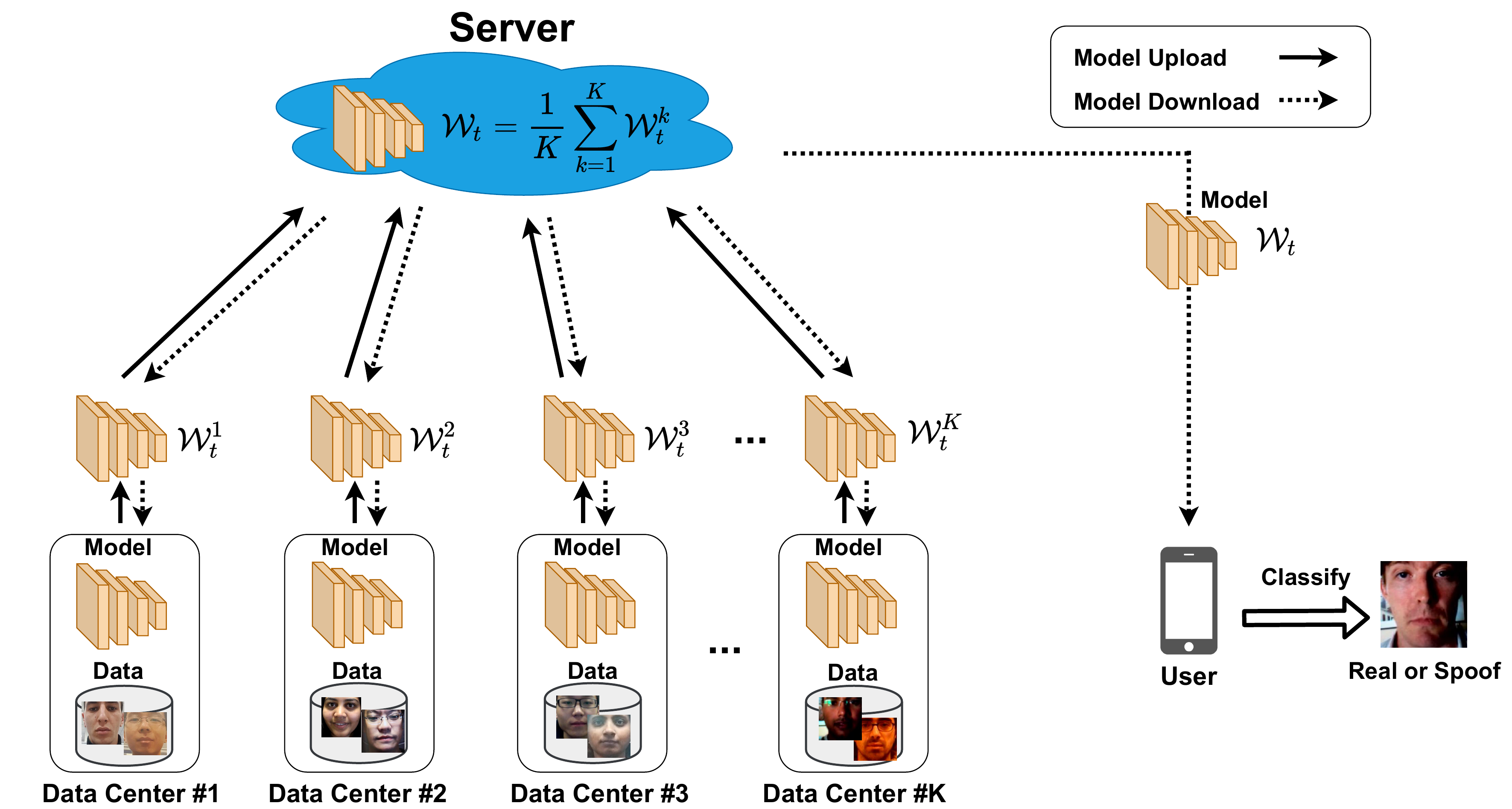}
		
	\end{center}
	
	\caption{Overview of the proposed FedPAD framework. Through several rounds of communication between \textit{data centers} and \textit{server}, the collaboratively trained global fPAD model parameterized by $\mathcal{W}_t$ can be obtained in a data privacy preserving way. Users can download this global model from the server to their device to detect various face presentation attacks.}
	\label{fig:Overview_FedPAD}
\end{figure}

\begin{algorithm}[t]
	\small
	\DontPrintSemicolon
	\caption{Federated Face Presentation Attack Detection}
	\label{algorithm_FedPAD}
	\SetAlgoLined
	\SetKwInOut{Input}{Input}
    \Input{ $K$ Data Centers have $K$ fPAD datasets $\mathcal{D}^{1}, \mathcal{D}^{2},..., \mathcal{D}^{K}$}

	\textbf{Server aggregates:} \\
	initialize $\mathcal{W}_0$ \\
	\For {\upshape each round $t$ = 0, 1, 2,... }{
		\For{\upshape each data center $k$ = 1, 2,..., $K$  \textbf{in parallel}}{
			${W}_t^k \leftarrow {\rm \textbf{DataCenterUpdate}}(k, \mathcal{W}_t)$
		}
		$\mathcal{W}_t=$ Eq.$(\ref{equ:fedavg})$ \\
		\textbf{Download} ${W}_t$ to \textbf{Data Centers}
	}
	\textbf{Users Download} ${W}_t$	\\
	\;
	\textbf{DataCenterUpdate}$(k, \mathcal{W})\textbf{:}$\\
	\For{\upshape each local epoch $1, 2,..., L$}{
	$\mathcal{L}(\mathcal{W}^k)=$ Eq.$(\ref{equ:fedpadcrossentropy}), \mathcal{W}^k \leftarrow \mathcal{W}^k-\eta \nabla\mathcal{L}(\mathcal{W}^k)$
	}
	\textbf{Upload} ${W}^k$ to \textbf{Server}
\end{algorithm}

\subsection{Federated Face Presentation Attack Detection}

The proposed FedPAD framework is summarized in Fig.~\ref{fig:Overview_FedPAD} and Algorithm~\ref{algorithm_FedPAD}. Suppose that $K$ \textit{data enters} collect their own fPAD datasets designed for different characteristics and scenarios of face presentation attacks. The corresponding collected fPAD datasets are denoted as $\mathcal{D}^{1}, \mathcal{D}^{2},..., \mathcal{D}^{K}$ with data provided with image and label pairs denoted as $x$ and $y$. $y$ are the ground-truth with binary class labels (y= 0/1 are labels of spoof/real samples). Based on the collected fPAD data, each \textit{data center} can train its own fPAD model by iteratively minimizing the cross-entropy loss as follows:

\begin{equation}
	\small
	\begin{split}
		\mathcal{L}(\mathcal{W}^k) = \sum\limits_{(x,y)\sim\mathcal{\mathcal{D}}^{k}}y\log\mathcal{F}^k(x)+(1-y)\log(1-\mathcal{F}^k(x)),
	\end{split}
	\label{equ:fedpadcrossentropy}
\end{equation} 
where the fPAD model $\mathcal{F}^{k}$ of the $k$-th \textit{data enter} is parameterized by $\mathcal{W}^k$ ($k=1,2,3,...,K$). After optimization with several local epochs via $$\mathcal{W}^k \leftarrow \mathcal{W}^k-\eta \nabla\mathcal{L}(\mathcal{W}^k),$$ each \textit{data enter} can obtain the trained fPAD model with the updated model parameters.

It should be noted that the dataset corresponding to each  \textit{data center} is from a specific input distribution and it only contains  a finite set of known types of spoof attack data. When a  model is trained using this data, it focuses on addressing the characteristics and scenarios of face presentation attacks prevalent in the corresponding dataset. However,  a model trained from a specific \textit{data center} will not generalize well to unseen face presentation attacks. It is a well-known fact that diverse fPAD training data contribute to a better generalized fPAD model. A straightforward solution is to collect and combine all the data from $K$ data centers denoted as $\mathcal{D} = \{\mathcal{D}^{1} \cup \mathcal{D}^{2}\cup...\cup \mathcal{D}^{K}\} $ to train a fPAD model. It has been shown that domain generalization and meta-learning based fPAD methods can further improve the generalization ability with the above combined multi-domain data $\mathcal{D}$~\cite{Shao_2019_CVPR,Shao_2020_AAAI}. However, when sharing data between different \textit{data centers} are prohibited due to the privacy issue, this naive solution is not practical.

To circumvent this limitation and enable various data centers to collaboratively train a fPAD model, we propose the FedPAD framework. Instead of accessing private fPAD data of each \textit{data center}, the proposed FedPAD framework introduces a \textit{server} to exploit the fPAD information of all data centers by aggregating the above model updates ($\mathcal{W}^1, \mathcal{W}^2,..., \mathcal{W}^K$) of all data centers. Inspired by the  Federated Averaging~\cite{mcmahan2016communication} algorithm, in the proposed framework, the server carries out the aggregation of model updates via calculating the average of updated parameters ($\mathcal{W}^1, \mathcal{W}^2,..., \mathcal{W}^K$) in all data centers as follows:
\begin{equation}
	\small
	\mathcal{W}=\frac{1}{K} \sum\limits_{k=1}^K\mathcal{W}^k.
\end{equation}
After the aggregation, the server produces a global fPAD model parameterized by $\mathcal{W}$ that exploits the fPAD information of various data centers without accessing the private fPAD data. We can further extend the above aggregation process into $t$ rounds. The server distributes the aggregated model $\mathcal{W}$ to every data center as the initial model parameters for the next-round updating of local parameters. Thus, data centers can obtain the $t$-th round updated parameters denoted as ($\mathcal{W}_t^1, \mathcal{W}_t^2,..., \mathcal{W}_t^K$). The $t$-th aggregation in the server can be carried out as follows:
\begin{equation}
	\small
	\mathcal{W}_t=\frac{1}{K} \sum\limits_{k=1}^K\mathcal{W}_t^k.
	\label{equ:fedavg}
\end{equation}
After  $t$-rounds of communication between data centers and the server, the trained global fPAD model parameterized by $\mathcal{W}_t$ can be obtained without compromising the private data of individual \textit{data centers}. Once training is converged, \textit{users} will download the trained model from the server to their devices to carry out fPAD locally.

\begin{figure*}[!htb]
	
	\begin{center}
		
		\includegraphics[width=0.9\linewidth]{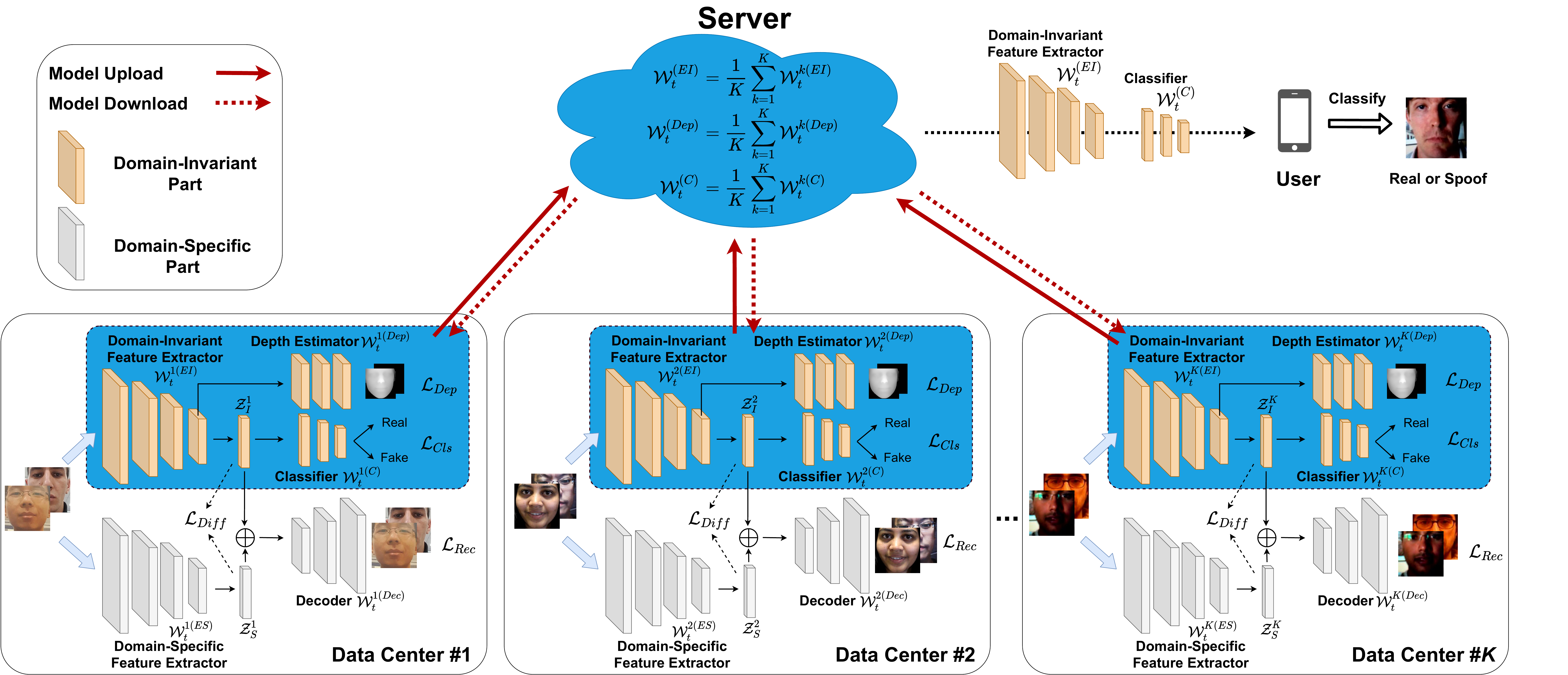}
		
	\end{center}
	
	\caption{Overview of the proposed FedGPAD framework. Through several rounds of communication between the domain-invariant parts of fPAD models of \textit{data centers} and \textit{server}, the collaboratively trained global fPAD model parameterized by $\mathcal{W}_t^{(EI)}, \mathcal{W}_t^{(Dep)}, \mathcal{W}_t^{(C)}$ can be obtained in a data privacy preserving way. Users can download this global feature extractor and classifier from the server to their device to detect various face presentation attacks.}
	\label{fig:Overview_FedGPAD}
\end{figure*}

\begin{algorithm}[t]
	\small
	\DontPrintSemicolon
	\caption{Federated Generalized Face Presentation Attack Detection}
	\label{algorithm_FedGPAD}
	\SetAlgoLined
	\SetKwInOut{Input}{Input}
	\Input{ $K$ Data Centers have $K$ fPAD datasets $\mathcal{D}^{1}, \mathcal{D}^{2},..., \mathcal{D}^{K}$}
	
	\textbf{Server aggregates:} \\
	initialize $\mathcal{W}_0^{(EI)},\mathcal{W}_0^{(C)},\mathcal{W}_0^{(Dep)}$ \\
	\For {\upshape each round $t$ = 0, 1, 2,... }{
		\For{\upshape each data center $k$ = 1, 2,..., $K$  \textbf{in parallel}}{
			$\mathcal{W}_t^{k(EI)},\mathcal{W}_t^{k(C)}, \mathcal{W}_t^{k(Dep)} \leftarrow $ \\
			${\rm \textbf{DataCenterUpdate}}(k, \mathcal{W}_t^{(EI)},\mathcal{W}_t^{(C)},\mathcal{W}_t^{(Dep)})$
		}
		$\mathcal{W}_t^{(EI)}, \mathcal{W}_t^{(C)}, \mathcal{W}_t^{(Dep)} =$ Eq.$(\ref{equ:fedGavg})$ \\
		\textbf{Download}$\mathcal{W}_t^{(EI)}, \mathcal{W}_t^{(C)}, \mathcal{W}_t^{(Dep)}$ to \textbf{Data Centers}
	}
	\textbf{Users Download} $\mathcal{W}_t^{(EI)}, \mathcal{W}_t^{(C)}$ \\
	\;
	\textbf{DataCenterUpdate}$(k, \mathcal{W}^{(EI)},\mathcal{W}^{(C)},\mathcal{W}^{(Dep)})\textbf{:}$\\
	$\mathcal{L}_{Cls}(\mathcal{W}^{k(EI)}, \mathcal{W}^{k(C)})=$ Eq.$(\ref{equ:crossentropyFedGPAD})$ \\
	$\mathcal{L}_{Dep}(\mathcal{W}^{k(EI)}, \mathcal{W}^{k(Dep)})=$ Eq.$(\ref{equ:depthFedGPAD})$ \\
	$\mathcal{L}_{Rec}(\mathcal{W}^{k(EI)}, \mathcal{W}^{k(ES)}, \mathcal{W}^{k(Dec)}) =$ Eq.$(\ref{equ:recGPAD})$ \\
	$\mathcal{L}_{Diff}(\mathcal{W}^{k(EI)}, \mathcal{W}^{k(ES)}) =$ Eq.$(\ref{equ:diffGPAD})$ \\
	$\mathcal{W}^{k(EI)},\mathcal{W}^{k(C)}, \mathcal{W}^{k(Dep)} \leftarrow $ Eq.$(\ref{equ:fedGavgOpt})$ \\
	\textbf{Upload} $\mathcal{W}^{k(EI)}, \mathcal{W}^{k(C)}, \mathcal{W}^{k(Dep)}$ to \textbf{Server}
\end{algorithm}

\subsection{Federated Generalized Face Presentation Attack Detection}

Although the above proposed FedPAD can effectively learn a fPAD model by exploiting the fPAD information available in all data centers in a privacy preserving manner, the generalization ability of the learned fPAD model is still limited and under-explored. Due to the different characteristics and scenarios of face presentation attacks in all data centers, the corresponding trained fPAD models will be easily guided to extract domain-specific features and thus the parameters of each fPAD model contain much domain-specific information. In the context of fPAD, these domain-specific features are usually related to display materials, background information or environmental noises~\cite{Shao_2019_CVPR,Shao_2020_AAAI}. These features will be changed if samples are presented with different display materials, or recorded under different environments with different backgrounds (listed in Table~\ref{tab:datasets}). Therefore, if directly aggregating such model parameters in the server, the aggregation will be severely affected by domain-specific information and thus the generalization ability of the aggregated fPAD model will be degraded.

To overcome this limitation of the FedPAD framework, we further propose an improved version called Federated Generalized Face Presentation Attack Detection (FedGPAD) framework, with the objective of exploiting a fPAD model with better generalization ability under the FL framework.   Fig.~\ref{fig:Overview_FedGPAD} and Algorithm~\ref{algorithm_FedGPAD} give an overview of FedGPAD.  Many existing domain generalization methods assume that by learning  domain-invariant features among multiple source domains, the model is more able to generalize well to the unseen target domain during testing~\cite{Shao_2019_CVPR}. Inspired by this idea, we treat every data center as a source domain and aim to learn a fPAD model focusing on extracting the domain-invariant features in the server. However, due to the privacy preserving issues required by FL, current domain-invariant feature learning-based domain generalization methods cannot be adapted as they require to get access to all source domain data from all local data centers simultaneously. To address this issue, we propose a novel \textbf{federated domain disentanglement} strategy in the FedGPAD framework, aiming to carry out domain-invariant feature learning in a privacy preserving way. Federated domain disentanglement composes of two iterative steps, \textit{i.e.}, \textbf{local domain disentanglement learning} and \textbf{domain-invariant model parameters communications}. Specifically, local domain disentanglement learning decomposes the fPAD model in each data center into domain-invariant part and domain-specific part, which disentangle the domain-invariant and domain-specific features from images, respectively. At this stage, the domain-invariant information extracted by domain invariant part of the fPAD model in each data center is still restricted by the corresponding local data distribution. To further learn domain-invariant information among all the data centers, every data center communicates with the server by uploading its domain-invariant part of the fPAD model to the server for model aggregation. In this way, the domain-invariant information from all the data centers will be summarized by the aggregated model parameters in the server and thus domain-invariant features among all the data centers can be learned by the aggregated model. In the next communication rounds, the aggregated domain-invariant model in the server will be downloaded to all the data centers, which will continually participate and optimize the local domain disentanglement learning in the local data centers. Therefore, the current domain-invariant feature learning is substituted by the iterative steps carried out in the federated domain disentanglement in a privacy preserving manner.

Based on the above idea, we decompose the $K$ fPAD models $\mathcal{F}^{k} (k=1,...,K)$  proposed in the FedPAD framework into the domain-invariant parts, namely $K$ Domain-Invariant Feature Extractors $EI^{k}$ parameterized by $\mathcal{W}^{k(EI)}$, $K$ Classifiers $C^{k}$ parameterized by $\mathcal{W}^{k(C)}$, and $K$ Depth Estimators $Dep^{k}$ parameterized by $\mathcal{W}^{k(Dep)}$; the domain-specific parts, namely $K$ Domain-Specific Feature Extractors $ES^{k}$ parameterized by $\mathcal{W}^{k(ES)}$ and $K$ Decoders $Dec^{k}$ parameterized by $\mathcal{W}^{k(Dec)}$.

As illustrated in Fig.~\ref{fig:Overview_FedGPAD}, we first carry out the first step: local domain disentanglement learning. In data center $k$, we try to disentangle domain-invariant features and domain-specific features by feeding the input data into domain-invariant feature extractor and domain-specific feature extractor, respectively: $\mathcal{Z}_{I}^{k}=EI^{k}(x)$, $\mathcal{Z}_{S}^{k}=ES^{k}(x)$. Since domain-invariant features are related to the classification of real and spoof faces, we train a domain-invariant fPAD model by minimizing the cross-entropy classification loss based on the domain-invariant features as follows:
\begin{equation}
	\small
	\begin{split}
		&\mathcal{L}_{Cls}(\mathcal{W}^{k(EI)}, \mathcal{W}^{k(C)}) \\
		&= \sum\limits_{(x,y)\sim\mathcal{\mathcal{D}}^{k}}y\log C^k(\mathcal{Z}_{I}^{k})+(1-y)\log(1-C^k(\mathcal{Z}_{I}^{k})).
	\end{split}
	\label{equ:crossentropyFedGPAD}
\end{equation} 

As mentioned in~\cite{Shao_2019_CVPR}, face depth information can provide more generalized liveness cues and thus aid in searching for the domain-invariant feature space. Inspired by this, we incorporate face depth map as the auxiliary supervision to regularize the domain-invariant feature learning with the depth estimation loss as follows:
\begin{equation}
	\small
	\begin{split}
		\mathcal{L}_{Dep}(\mathcal{W}^{k(EI)}, \mathcal{W}^{k(Dep)}) = \sum\limits_{(x,M)\sim\mathcal{\mathcal{D}}^{k}} \left\| Dep^{k}(\mathcal{Z}_{I}^{k})-M \right\|_2^2,
	\end{split}
	\label{equ:depthFedGPAD}
\end{equation} 
where $M$ corresponds to pseudo face depth map used as the ground-truth for live face images and zero map for spoof images. 

Domain-invariant features combined with domain-specific features should encode the complete features from the input data and they should complement each other. Therefore, we should well reconstruct the input data by decoding the combined domain-invariant and domain-specific features via a decoder. This can be realized by minimizing the reconstruction loss between the input data and the reconstructed data as follows:
\begin{equation}
	\small
	\begin{split}
		&\mathcal{L}_{Rec}(\mathcal{W}^{k(EI)}, \mathcal{W}^{k(ES)}, \mathcal{W}^{k(Dec)}) \\
		& = \sum\limits_{x\sim\mathcal{\mathcal{D}}^{k}} \left\| Dec^{k}(\mathcal{Z}_{I}^{k}+\mathcal{Z}_{S}^{k})-x \right\|_2^2.
	\end{split}
	\label{equ:recGPAD}
\end{equation} 

Moreover, since domain-invariant and domain-specific encoders should encode different aspects of the input data, 
%inspired by~\cite{bousmalis2016domain}, 
to better disentangle the two types of features, we further impose a soft subspace orthogonality constraint between the domain-invariant and domain-specific features via a difference loss as follows:
\begin{equation}
	\small
	\begin{split}
		\mathcal{L}_{Diff}(\mathcal{W}^{k(EI)}, \mathcal{W}^{k(ES)}) = \sum\limits_{x\sim\mathcal{\mathcal{D}}^{k}} \left\| (\mathcal{Z}_{I}^{k})^T (\mathcal{Z}_{S}^{k})\right\|_F^2,
	\end{split}
	\label{equ:diffGPAD}
\end{equation} 
where $\left\|\cdot \right\|_F^2$ is the squared Frobenius norm. The difference loss encourages orthogonality between the domain-invariant and domain-specific features and thus they can be better disentangled.

The overall loss function of the local domain disentanglement learning in local data center $k$ is as follows:
\begin{equation}
	\begin{split}
		&\mathcal{L}(\mathcal{W}^{k(EI)}, \mathcal{W}^{k(ES)}, \mathcal{W}^{k(Dep)}, \mathcal{W}^{k(C)}, \mathcal{W}^{k(Dec)}) = \\
		&\mathcal{L}_{Cls}(\mathcal{W}^{k(EI)}, \mathcal{W}^{k(C)}) + \mathcal{L}_{Dep}(\mathcal{W}^{k(EI)}, \mathcal{W}^{k(Dep)})\\
		&\mathcal{L}_{Rec}(\mathcal{W}^{k(EI)}, \mathcal{W}^{k(ES)}, \mathcal{W}^{k(Dec)})+	\mathcal{L}_{Diff}(\mathcal{W}^{k(EI)}, \mathcal{W}^{k(ES)}).
	\end{split}
\label{equ:loss}
\end{equation} 

Note that the domain-invariant part (i.e. domain-invariant feature extractors, classifiers, and depth estimators) of fPAD models are purified with domain-invariant information. Therefore, we can carry out the second step -- domain-invariant model parameters communication.  That is, we communicate the parameters of the domain-invariant part of fPAD model in every local data center with the server. Specifically, at the communication round $t$, we update model parameters of the domain-invariant part of the fPAD models after the local domain disentanglement learning as follows:
\begin{equation}
	\small
	\begin{split}
		&\mathcal{W}_t^{k(EI)} \leftarrow \mathcal{W}_t^{k(EI)} \\
		&-\eta \nabla\mathcal(\mathcal{L}(\mathcal{W}_t^{k(EI)}, \mathcal{W}_t^{k(ES)}, \mathcal{W}_t^{k(Dep)}, \mathcal{W}_t^{k(C)}, \mathcal{W}_t^{k(Dec)})) \\
		&\mathcal{W}_t^{k(C)} \leftarrow \mathcal{W}_t^{k(C)} -\eta \nabla\mathcal({L}_{Cls}(\mathcal{W}_t^{k(EI)}, \mathcal{W}_t^{k(C)}))\\
		&\mathcal{W}_t^{k(Dep)} \leftarrow \mathcal{W}_t^{k(Dep)} -\eta \nabla\mathcal(\mathcal{L}_{Dep}(\mathcal{W}_t^{k(EI)}, \mathcal{W}_t^{k(Dep)})).
	\end{split}
	\label{equ:fedGavgOpt}
\end{equation} 
We upload the above updated model parameters of the domain-invariant part to the server for model parameter aggregation by calculating the average of the updated parameters in all data centers as follows:
\begin{equation}
	\small
	\begin{split}
		&\mathcal{W}_t^{(EI)}=\frac{1}{K} \sum\limits_{k=1}^K\mathcal{W}_t^{k(EI)}, \\
		&\mathcal{W}_t^{(C)}=\frac{1}{K} \sum\limits_{k=1}^K\mathcal{W}_t^{k(C)}, \\
		&\mathcal{W}_t^{(Dep)}=\frac{1}{K} \sum\limits_{k=1}^K\mathcal{W}_t^{k(Dep)}.
	\end{split}
	\label{equ:fedGavg}
\end{equation}
The server downloads these aggregated domain-invariant parts of the fPAD models $\mathcal{W}_t^{(EI)}, \mathcal{W}_t^{(C)}, \mathcal{W}_t^{(Dep)}$ to every data center as the initial model parameters for the next-round of local model optimization. The downloaded domain-invariant parts of the fPAD models contain the updated domain-invariant information, which can reversely participate and optimize the local domain disentanglement learning carried out in each local data center. 

After  $t$-rounds of communications between data centers and the server, the trained global fPAD model parameterized by $\mathcal{W}_t^{(EI)}, \mathcal{W}_t^{(C)}, \mathcal{W}_t^{(Dep)}$ can be obtained without compromising the private data of the individual \textit{data centers}. Once training is converged, \textit{users} will download the trained domain-invariant feature extractor and classifier $\mathcal{W}_t^{(EI)}, \mathcal{W}_t^{(C)}$ from the server to their devices to carry out fPAD locally.

\section{Experiments}
\begin{figure*}[!htb]
	\begin{center}
		\includegraphics[height=2.8cm, width=0.9\linewidth]{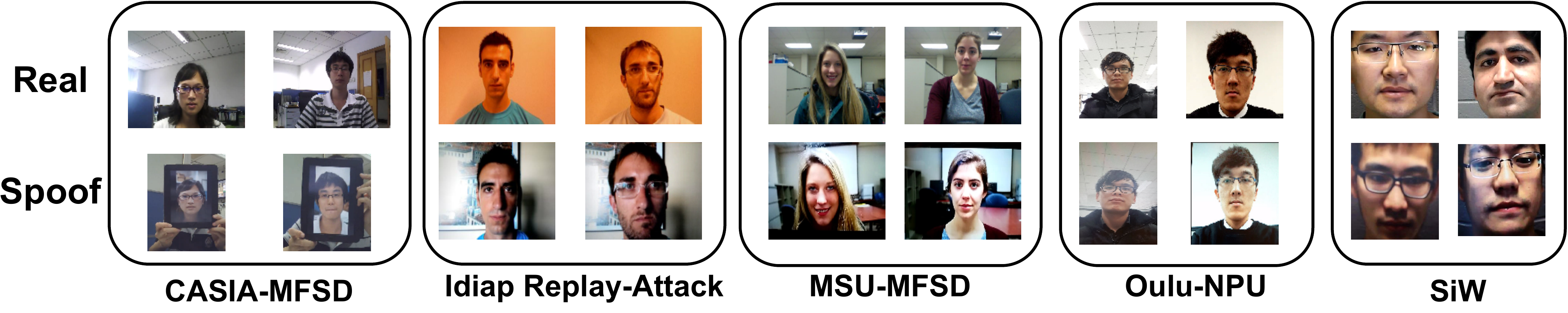}
	\end{center}
	\caption{Sample images corresponding to real and attacked faces from CASIA-MFSD~\cite{2012ICBcasia}, Idiap Replay-Attack~\cite{2012BIOSIGidiap}, MSU-MFSD~\cite{2015TIFSida}, Oulu-NPU~\cite{2017FGoulu}, SiW~\cite{2018CVPRauxliary} datasets.}
	\label{fig:datasets}
\end{figure*}

This paper focuses on exploiting FL to deal with 2-dimensional (2D) face presentation attacks. To evaluate the performance of the proposed FL frameworks, we carry out experiments using five 2D fPAD datasets. In this section, we first describe the datasets, testing protocols evaluation metrics and implementation details. Then we report various experimental results based on multiple fPAD datasets. Discussions and analysis about the results are provided along with various insights about FL for fPAD.

\subsection{Experimental Settings}

\subsubsection{Datasets}

\begin{table}[htb]
	\renewcommand{\arraystretch}{1}
	\centering	
	\scriptsize
	\caption{Comparison of five experimental datasets.}	
	\begin{tabular}{c|c|c|c|c}
	\toprule[1pt]
		\textbf{Dataset} & \begin{tabular}[c]{@{}c@{}}\textbf{Extra }\\\textbf{ light}\end{tabular} & \begin{tabular}[c]{@{}c@{}}\textbf{Complex}\\ \textbf{background}\end{tabular} & \begin{tabular}[c]{@{}c@{}}\textbf{Attack}\\ \textbf{type}\end{tabular}                                  & \begin{tabular}[c]{@{}c@{}}\textbf{Display} \\ \textbf{devices}\end{tabular}           \\ \hline
		C       & No                                                     & Yes                                                           & \begin{tabular}[c]{@{}c@{}}Printed photo\\ Cut photo\\ Replayed video\end{tabular}     & iPad                                                                 \\ \hline
		I       & Yes                                                    & Yes                                                          & \begin{tabular}[c]{@{}c@{}}Printed photo\\ Display photo\\ Replayed video\end{tabular} & \begin{tabular}[c]{@{}c@{}}iPhone 3GS \\ iPad\end{tabular}           \\ \hline
		M       & No                                                     & Yes                                                          & \begin{tabular}[c]{@{}c@{}}Printed photo\\ Replayed video\end{tabular}                 & \begin{tabular}[c]{@{}c@{}}iPad Air\\ iPhone 5S\end{tabular}         \\ \hline
		O       & Yes                                                    & No                                                           & \begin{tabular}[c]{@{}c@{}}Printed photo\\ Display photo\\ Replayed video\end{tabular} & \begin{tabular}[c]{@{}c@{}}Dell 1905FP\\ Macbook Retina\end{tabular} \\ \hline
		
		S       & Yes                                                    & Yes	                                                           & \begin{tabular}[c]{@{}c@{}}Printed photo\\ Display photo\\ Replayed video\end{tabular} & \begin{tabular}[c]{@{}c@{}}Dell 1905FP\\iPad Pro\\iPhone 7\\Galaxy S8\\ Asus MB168B\end{tabular} \\
	\bottomrule[1pt]
	\end{tabular}
	\label{tab:datasets}
\end{table}

We evaluate our method using the following five publicly available 2D fPAD datasets which contain print, video-replay attacks:\\
1) Oulu-NPU~\cite{2017FGoulu} (O for short)\\
2) CASIA-MFSD~\cite{2012ICBcasia} (C for short)\\ 
3) Idiap Replay-Attack~\cite{2012BIOSIGidiap} (I for short)\\ 
4) MSU-MFSD~\cite{2015TIFSida} (M for short)\\
5) SiW~\cite{2018CVPRauxliary} (S for short)\\
%6) 3DMAD~\cite{Marceltifs3D2014} (3 for short)\\
%7) HKBUMARsV2~\cite{2018ECCVrPPG} (H for short).\\ 
The OULU-NPU dataset~\cite{2017FGoulu} is composed of 4950 real access and attack videos, which are displayed with Dell UltraSharp 1905FP display with 1280 $\times$ 1024 resolution and 2015 Macbook 13” laptop with Retina display of 2560 $\times$ 1600 resolution. We extract total 42248 frames from videos in the OULU-NPU dataset for experiments. The CASIA-MFSD dataset~\cite{2012ICBcasia} consists of 600 videos with various imaging qualities: low, normal, and high. We collect in total 54126 frames from videos in the CASIA-MFSD dataset for evaluation. The Idiap Replay-Attack dataset~\cite{2012BIOSIGidiap} contains 1200 videos displayed on the screen of an iPhone 3GS and on the screen of an iPad with a resolution of 1024 $\times$ 168. 59000 frames from videos in the Idiap Replay-Attack dataset are extracted for experiments. The MSU-MFSD dataset~\cite{2015TIFSida}  consists of 440 videos which are replayed on an iPad Air screen and iPhone 5S screen with a resolution of 1920 $\times$ 1080. Totally 38891 frames from videos in the MSU-MFSD dataset are captured for evaluation.
The SiW dataset~\cite{2018CVPRauxliary} provides 4620 videos displayed with 2 Print and 4 Replay devices (summarized in Table~\ref{tab:datasets} in manuscript). We extract in total 89540 frames from videos in the SiW dataset for experiments.

Table~\ref{tab:datasets} shows the variations in these datasets. Sample images from these datasets are shown in Fig.~\ref{fig:datasets}. From Table~\ref{tab:datasets} and Fig.~\ref{fig:datasets} it can be seen that different fPAD datasets exploit different characteristics and scenarios of face presentation attacks (\textit{i.e.}, different attack types, display materials and resolution, illumination, background and so on). Therefore, significant domain shifts exist among these datasets.

\begin{table}[!htb]
	\renewcommand{\arraystretch}{1}
	\centering
	\scriptsize
	\caption{Details of the proposed network in the FedGPAD framework.}
	\begin{tabular}{ccc}
		\toprule[1pt]
		\textbf{Layer}             & \textbf{Chan./Stri.}            & \textbf{Out.Size}            \\ \hline\hline
		\multicolumn{3}{c}{\textbf{Domain-invariant/specific Feature Extractor}}                    \\ \hline\hline
		\multicolumn{3}{c}{\begin{tabular}[c]{@{}c@{}}Input\\ image\end{tabular}}                   \\ \hline
		conv1-1                    & 64/1                            & 256                          \\
		conv1-2                    & 128/1                           & 256                          \\
		conv1-3                    & 196/1                           & 256                          \\
		conv1-4                    & 128/1                           & 256                          \\
		pool1-1                    & -/2                             & 128                          \\
		conv1-5                    & 128/1                           & 128                          \\
		conv1-6                    & 196/1                           & 128                          \\
		conv1-7                    & 128/1                           & 128                          \\
		pool1-2                    & -/2                             & 64                           \\
		conv1-8                    & 128/1                           & 64                           \\
		conv1-9                    & 196/1                           & 64                           \\
		conv1-10                   & 128/1                           & 64                           \\
		pool1-3                    & -/2                             & 32                           \\
		conv1-11                   & 128/1                           & 32                           \\
		pool1-4                    & -/2                             & 16                           \\
		conv1-12                   & 256/1                           & 16                           \\
		pool1-5                    & -/2                             & 8                            \\
		conv1-13                   & 512/1                           & 8                            \\ \hline\hline
		\multicolumn{3}{c}{\textbf{Decoder}}                                                        \\ \hline\hline
		\multicolumn{3}{c}{\begin{tabular}[c]{@{}c@{}}Input\\ conv1-13\end{tabular}}                \\ \hline
		\multicolumn{3}{c}{Average pooling}                                                         \\
		fc2-1                      & 1/1                             & 512*4*4                      \\
		deconv2-1                  & 512/2                           & 8                            \\
		deconv2-2                  & 256/2                           & 16                           \\
		deconv2-3                  & 128/2                           & 32                           \\
		deconv2-4                  & 64/2                            & 64                           \\
		deconv2-5                  & 32/2                            & 128                          \\
		deconv2-6                  & 6/2                             & 256                          \\ \hline\hline
		\multicolumn{3}{c}{\textbf{Classifier}}                                                     \\ \hline\hline
		\multicolumn{3}{c}{\begin{tabular}[c]{@{}c@{}}Input\\ conv1-13\end{tabular}}                \\ \hline
		\multicolumn{3}{c}{Average pooling}                                                         \\
		fc3-1                      & 1/1                             & 512                          \\
		fc3-2                      & 1/1                             & 1                            \\ \hline\hline
		\multicolumn{3}{c}{\textbf{Depth Estimator}}                                                \\ \hline\hline
		\multicolumn{3}{c}{\begin{tabular}[c]{@{}c@{}}Input\\ pool1-1+pool1-2+pool1-3\end{tabular}} \\ \hline
		conv4-1                    & 128/1                           & 32                           \\
		conv4-2                    & 64/1                            & 32                           \\
		conv4-3                    & 1/1                             & 32                          \\
	\bottomrule[1pt]
	\end{tabular}
	\label{tab:netFedGPAD}
\end{table}

\subsubsection{Protocol} 
The testing protocol used in the paper is designed to evaluate the generalization ability of fPAD models under the FL framework. Therefore, the performance of a trained model is evaluated against a dataset that has not been observed during training. In particular, we choose one dataset at a time to emulate the role of \textit{users} and consider all other datasets as \textit{data centers}. Real images and spoof images of \textit{data centers} are used to train a fPAD model. The trained model is tested considering the dataset that emulates the role of \textit{users}. We evaluate the performance of the model by considering how well the model is able to differentiate between real and spoof images belonging to each \textit{user}.

\subsubsection{Evaluation Metrics} 
Half Total Error Rate (HTER)~\cite{2004HTER} (half of the summation of false acceptance rate and false rejection rate),  Equal Error Rates (EER) and Area Under Curve (AUC) are used as evaluation metrics in our experiments, which are three most widely-used metrics for the cross-datasets/cross-domain evaluations. Following ~\cite{liu2019deep}, in the absence of a development set,  thresholds required for calculating evaluation metrics are determined based on the data in all \textit{data centers}.

\subsubsection{Implementation Details} 
Our deep network is implemented on the platform of PyTorch. For the network in the FedPAD framework, we directly adopt Resnet-18~\cite{Kaiming_Resnet_CVPR2016} as the structure of fPAD models. For the network in the FedGPAD framework, the structure details of domain-invariant/specific feature extractor, decoder, classifier, and depth estimator are shown in Table~\ref{tab:netFedGPAD}. Each convolutional layer in the feature extractor, classifier and depth estimator is followed by a batch normalization layer and a rectified linear unit (ReLU) activation function, with a kernel size of 3$\times$3. Each deconvolutional layer in the decoder is followed by a batch normalization layer and a LeakyReLU activation function, with kernel size 4$\times$4. The size of input image is 256$\times$256$\times$6, where we extract the RGB and HSV channels of each input image. Adam optimizer~\cite{adam} is used for the optimization. The learning rate is set as 1e-2. The batch size is 64 per data center for FedPAD and 16 for FedGPAD.

\begin{table*}[htb]
	\renewcommand{\arraystretch}{1}
	\centering	
	\caption{Comparison with models trained by data from single data center and various data centers.}
	\begin{tabular}{c|c|c|c|c|c|c|c|c}
		\toprule[1pt]
		\textbf{Methods}                  & \textbf{Data Centers} & \textbf{User} & \textbf{HTER (\%)} & \textbf{EER (\%)} & \textbf{AUC (\%)} & \textbf{Avg. HTER}          & \textbf{Avg. EER}          & \textbf{Avg. AUC}           \\ \hline
		\multirow{12}{*}{\textbf{Single}} & O                     & M             & 41.29              & 37.42             & 67.93             & \multirow{12}{*}{36.43} & \multirow{12}{*}{34.31} & \multirow{12}{*}{70.36} \\
		& C                     & M             & 27.09              & 24.69             & 82.91             &                         &                         &                         \\ 
		& I                     & M             & 49.05              & 20.04             & 85.89             &                         &                         &                         \\ 
		& O                     & C             & 31.33              & 34.73             & 73.19             &                         &                         &                         \\ 
		& M                     & C             & 39.80              & 40.67             & 66.58             &                         &                         &                         \\ 
		& I                     & C             & 49.25              & 47.11             & 55.41             &                         &                         &                         \\ 
		& O                     & I             & 42.21              & 43.05             & 54.16             &                         &                         &                         \\ 
		& C                     & I             & 45.99              & 48.55             & 51.24             &                         &                         &                         \\ 
		& M                     & I             & 48.50              & 33.70             & 66.29             &                         &                         &                         \\ 
		& M                     & O             & 29.80              & 24.12             & 84.86             &                         &                         &                         \\ 
		& C                     & O             & 33.97              & 21.24             & 84.33             &                         &                         &                         \\ 
		& I                     & O             & 46.95              & 35.16             & 71.58             &                         &                         &                         \\ \hline
		\multirow{4}{*}{\textbf{Fused}}    & O\&C\&I               & M             & 34.42              & 23.26             & 81.67             & \multirow{4}{*}{35.75}  & \multirow{4}{*}{31.29}  & \multirow{4}{*}{73.89}  \\ 
		& O\&M\&I               & C             & 38.32              & 38.31             & 67.93            &                         &                         &                         \\ 
		& O\&C\&M               & I             & 42.21              & 41.36             & 59.72             &                         &                         &                         \\ 
		& I\&C\&M               & O             & 28.04              & 22.24             & 86.24             &                         &                         &                         \\ \hline
		\multirow{4}{*}{\textbf{FedPAD}}    & O\&C\&I               & M             & 19.45              & 17.43             & 90.24             & \multirow{4}{*}{32.17}  & \multirow{4}{*}{28.84}  & \multirow{4}{*}{76.51}  \\ 
		& O\&M\&I               & C             & 42.27              & 36.95             & 70.49             &                         &                         &                         \\ 
		& O\&C\&M               & I             & 32.53              & 26.54             & 73.58             &                         &                         &                         \\ 
		& I\&C\&M               & O             & 34.44              & 34.45             & 71.74             &                         &                         &                         \\ \hline
		\multirow{4}{*}{\textbf{FedGPAD}}    & O\&C\&I               & M             & \textbf{12.73}              & \textbf{13.36}             & \textbf{91.25}             & \multirow{4}{*}{\textbf{18.59}}  & \multirow{4}{*}{\textbf{17.48}}  & \multirow{4}{*}{\textbf{89.25}}  \\ 
		& O\&M\&I               & C             & \textbf{28.69}              & \textbf{27.55}             & \textbf{80.58}             &                         &                         &                         \\ 
		& O\&C\&M               & I             & \textbf{10.97}              & \textbf{11.11}             & \textbf{95.34}             &                         &                         &                         \\ 
		& I\&C\&M               & O             & \textbf{21.95}              & \textbf{17.91}             & \textbf{89.85}             &                         &                         &                         \\ \hline
		\hline			
		\multirow{4}{*}{\begin{tabular}[c]{@{}c@{}}\textbf{All}\end{tabular}}     & O\&C\&I               & M             & 21.80              & 17.18             & 90.96             & \multirow{4}{*}{27.26}  & \multirow{4}{*}{25.09}  & \multirow{4}{*}{80.42}  \\ 
		& O\&M\&I               & C             & 29.46              & 31.54             & 76.29             &                         &                         &                         \\ 
		& O\&C\&M               & I             & 30.57              & 25.71             & 72.21             &                         &                         &                         \\ 
		& I\&C\&M               & O             & 27.22              & 25.91             & 82.21             &                         &                         &                         \\
	\bottomrule[1pt]
	\end{tabular}
	\label{tab:singleallours}
\end{table*}

\subsection{Experimental Results}

\begin{figure*}[htb]
	\centering
	\begin{minipage}[t]{0.24\linewidth}
		\centering
		\includegraphics[width=1\linewidth]{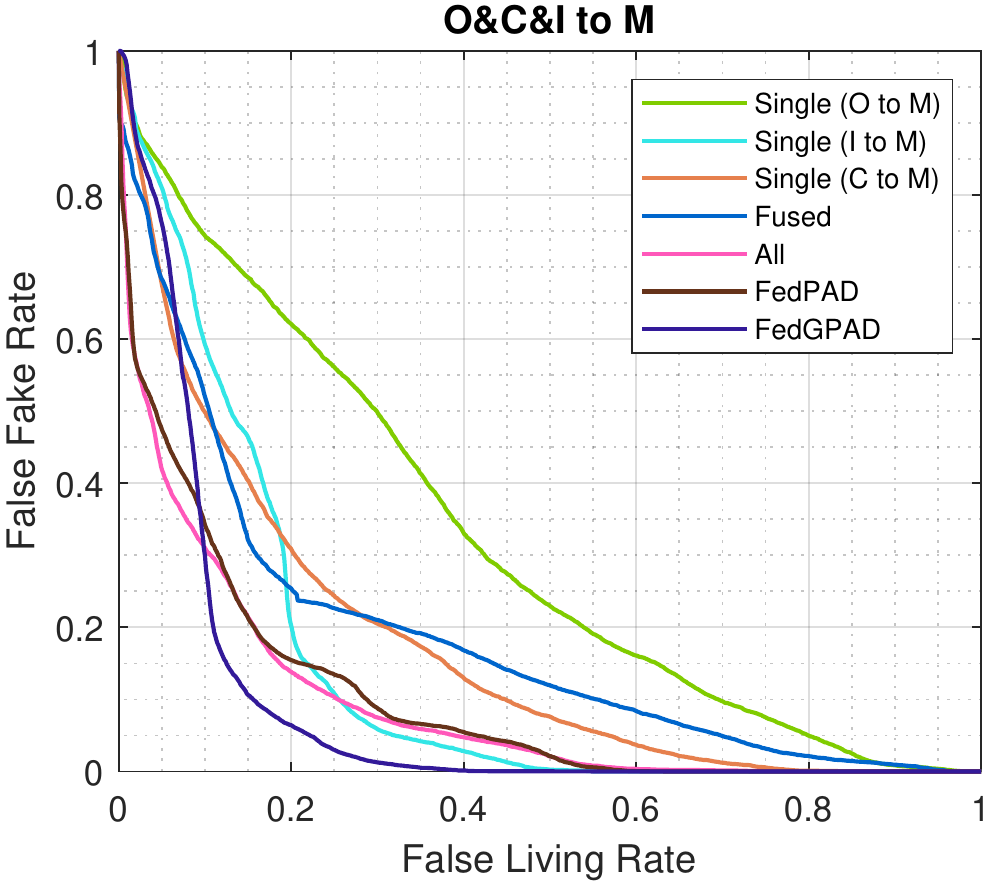}
	\end{minipage}%
	\begin{minipage}[t]{0.24\linewidth}
		\centering
		\includegraphics[width=1\linewidth]{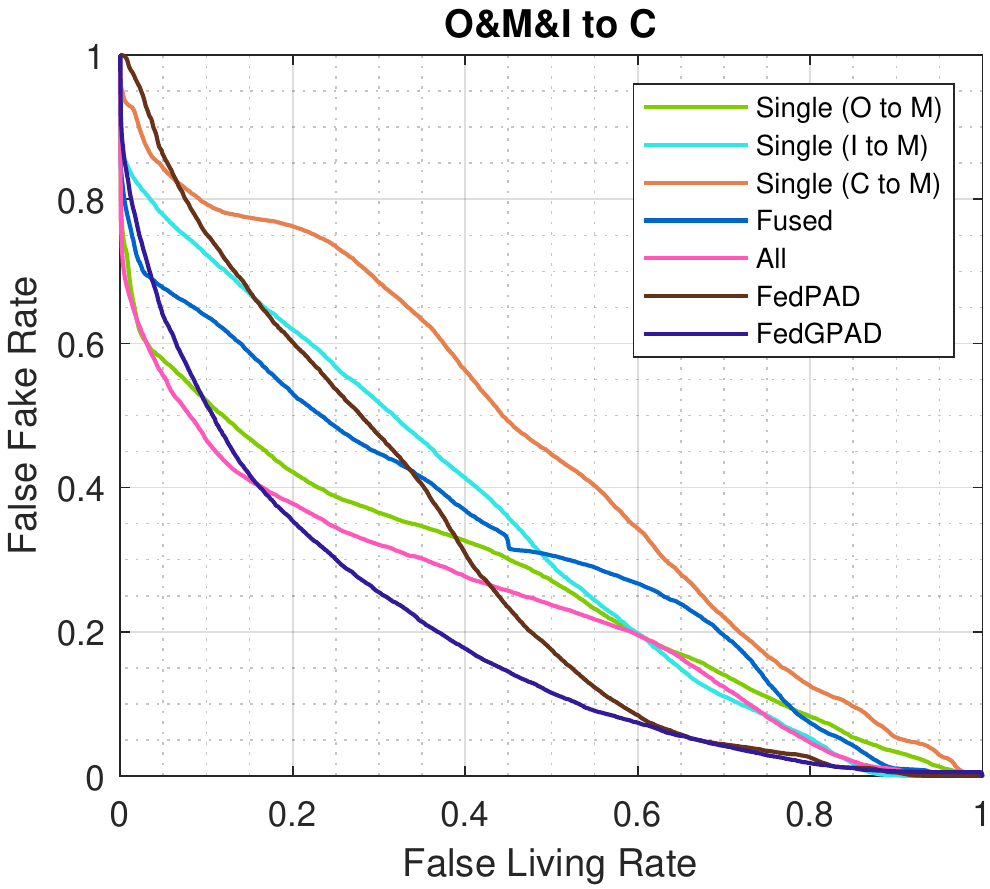}  
	\end{minipage}
	\begin{minipage}[t]{0.24\linewidth}
		\centering
		\includegraphics[width=1\linewidth]{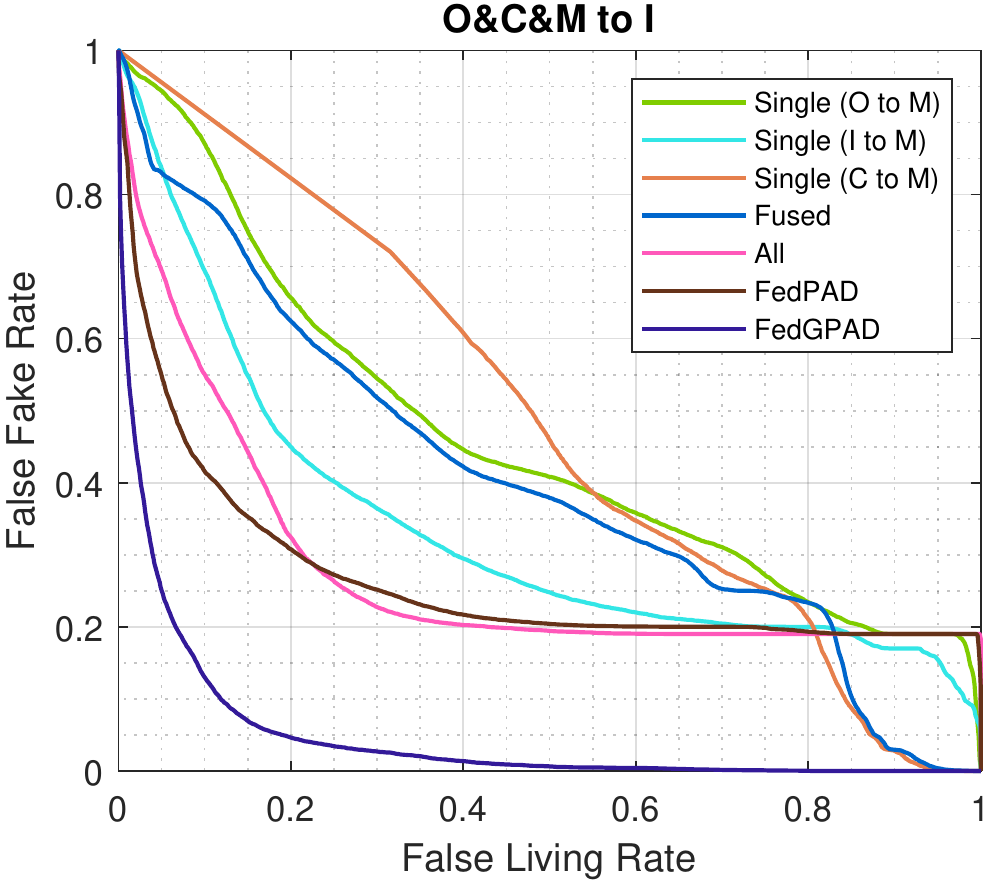}
	\end{minipage}%
	\begin{minipage}[t]{0.24\linewidth}
		\centering
		\includegraphics[width=1\linewidth]{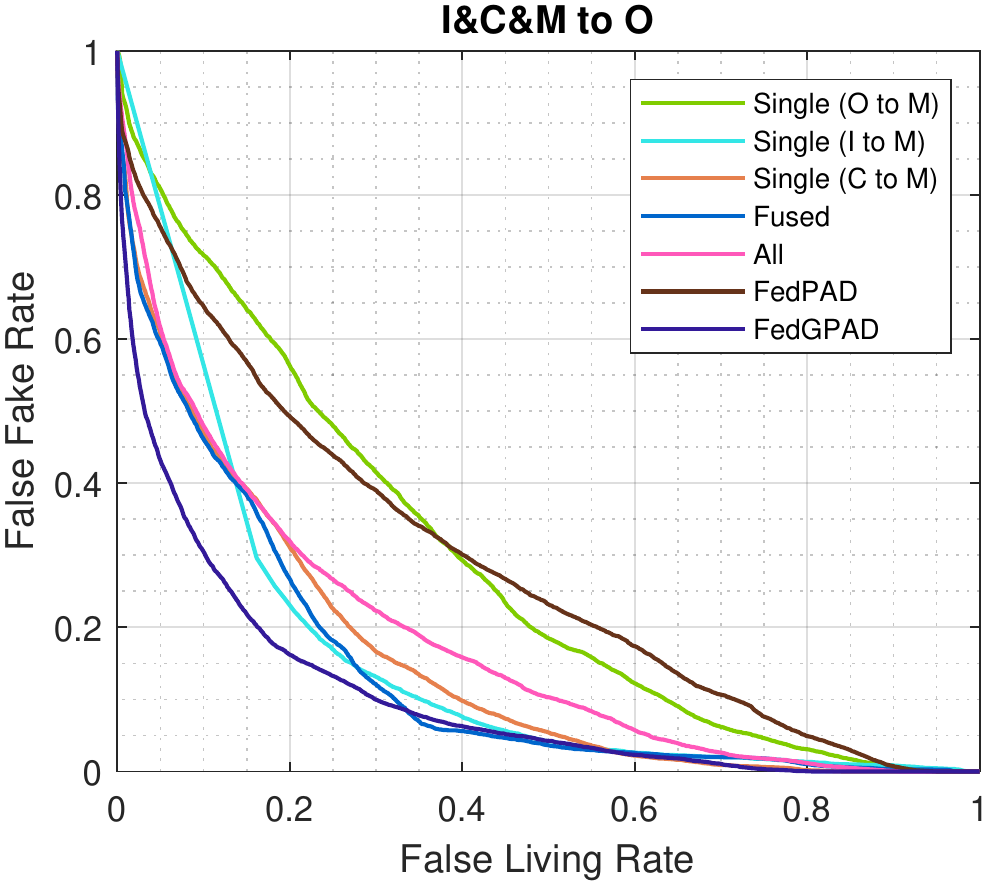}  
	\end{minipage}
	\caption{ROC curves of models trained by data from single data center and various data centers.}
	\label{fig:rocmain}
\end{figure*}

\begin{table*}[htb]
	\renewcommand{\arraystretch}{1}
	\centering	
	\caption{Comparison (\%) to face presentation attack detection methods on four testing sets for domain generalization on face presentation attack detection, e.g. O\&C\&I to M means data centers are O, C and I, and data from dataset M is presented to users.}
	\begin{tabular}{c|p{1cm}<{\centering}|p{1cm}<{\centering}|p{1cm}<{\centering}|p{1cm}<{\centering}|p{1cm}<{\centering}|p{1cm}<{\centering}|p{1cm}<{\centering}|p{1cm}<{\centering}|p{1cm}<{\centering}|p{1cm}<{\centering}}
		\toprule[1pt] 
		\multirow{2}{*}{\textbf{Method}} & \multicolumn{2}{c|}{\textbf{O\&C\&I to M}} & \multicolumn{2}{c|}{\textbf{O\&M\&I to C}} & \multicolumn{2}{c|}{\textbf{O\&C\&M to I}} & \multicolumn{2}{c|}{\textbf{I\&C\&M to O}} & \multicolumn{2}{c}{\textbf{Avg.}} \\ \cline{2-11} 
		& HTER         & AUC        & HTER         & AUC        & HTER         & AUC        & HTER         & AUC        & HTER     & AUC    \\ \hline
		& \multicolumn{10}{c}{\textbf{Without Considering Privacy Issue}}                                                                                                                   \\ \hline
		MS\_LBP~\cite{2011IJCBmstexture}                 & 29.76            & 78.50          & 54.28            & 44.98          & 50.30            & 51.64          & 50.29            & 49.31          & 46.15        & 56.10      \\
		Binary CNN~\cite{2014arxivdeepfeat}             & 29.25            & 82.87          & 34.88            & 71.94          & 34.47            & 65.88          & 29.61            & 77.54          & 32.05        & 74.55      \\
		IDA ~\cite{2015TIFSida}                    & 66.67            & 27.86          & 55.17            & 39.05          & 28.35            & 78.25          & 54.20            & 44.59          & 51.09        & 47.43      \\
		Color Texture ~\cite{2016TIFScolortxt}          & 28.09            & 78.47          & 30.58            & 76.89          & 40.40            & 62.78          & 63.59            & 32.71          & 40.66        & 62.71      \\
		LBPTOP ~\cite{2014EJIVPlbptop}                 & 36.90            & 70.80          & 42.60            & 61.05          & 49.45            & 49.54          & 53.15            & 44.09          & 45.52        & 56.37      \\
		Auxiliary(Depth Only)~\cite{2018CVPRauxliary}   & 22.72            & 85.88          & 33.52            & 73.15          & 29.14            & 71.69          & 30.17            & 77.61          & 28.88        & 77.08      \\
		MMD-AAE~\cite{2018CVPRdgadv}                & 27.08            & 83.19          & 44.59            & 58.29          & 31.58            & 75.18          & 40.98            & 63.08          & 36.05        & 69.93      \\
		MADDG~\cite{Shao_2019_CVPR}                   & 17.69            & 88.06          & 24.50            & 84.51          & 22.19            & 84.99          & 27.98            & 80.02          & 23.09        & 84.39      \\
		DR-MD-Net~\cite{wang2020cross}               & 17.02            & 90.10          & 19.68            & 87.43          & 20.87            & 86.72          & 25.02            & 81.47          & 20.64        & 86.43      \\
		RFMeta~\cite{Shao_2020_AAAI}                  & 13.89            & 93.98          & 20.27            & 88.16          & 17.30            & 90.48          & 16.45            & 91.16          & 16.97        & 90.94      \\
		NAS-Baseline~\cite{yu2020fas}            & 14.63            & 94.26          & 17.24            & 87.48          & 19.73            & 88.52          & 19.81            & 86.80          & 17.85        & 89.26      \\
		NAS-Baseline w/ D-Meta~\cite{yu2020fas}  & 11.62            & 95.85          & 16.96            & 89.73          & 16.82            & 91.68          & 18.64            & 88.45          & 16.01        & 91.42      \\
		NAS-FAS ~\cite{yu2020fas}                & 19.53            & 88.63          & 16.54            & 90.18          & 14.51            & 93.84          & 13.80            & 93.43          & 16.09        & 91.52      \\
		NAS-FAS w/ D-Meta~\cite{yu2020fas}       & 16.85            & 90.42          & 15.21            & 92.64          & 11.63            & 96.98          & 13.16            & 94.18          & 14.21        & 93.55 \\    
		DC-CDN~\cite{yu2021dual}      	& 25.51           & 81.80          & 15.00           & 92.80        & 15.88            & 91.61         & 18.82            & 89.86          & 18.80      & 89.01 \\ \hline
		& \multicolumn{10}{c}{\textbf{Considering Privacy Issue}}                                                                                                                           \\ \hline
		\textbf{FedPAD}    & 19.45          & 90.24         & 42.27           & 70.49         & 32.53            & 73.58         & 34.44            & 71.74          & 32.17        & 76.51  \\
		\textbf{FedGPAD}                 & 12.73            & 91.25          & 28.69            & 80.58          & 10.97            & 95.34          & 21.95            & 89.85          & 18.59        & 89.25      \\ 
	 \bottomrule[1pt]
	\end{tabular}
	\label{tab:STOA}
\end{table*}

\begin{table*}[htb]
	\renewcommand{\arraystretch}{1}
	\centering	
	\caption{Comparison (\%) to federated learning methods on four testing sets for domain generalization on face presentation attack detection.}
	\begin{tabular}{c|c|c|c|c|c|c|c|c|c|c|c|c|c|c|c}
		\toprule[1pt]
		\multirow{2}{*}{\textbf{Method}} & \multicolumn{3}{c|}{\textbf{O\&C\&I to M}} & \multicolumn{3}{c|}{\textbf{O\&M\&I to C}} & \multicolumn{3}{c|}{\textbf{O\&C\&M to I}} & \multicolumn{3}{c|}{\textbf{I\&C\&M to O}} & \multicolumn{3}{c}{\textbf{Avg.}} \\ \cline{2-16} 
		& HTER & EER       & AUC        & HTER     & EER    & AUC        & HTER   & EER     & AUC        & HTER    & EER    & AUC        & HTER    & EER   & AUC    \\ \hline
		FedAvg~\cite{mcmahan2016communication}       & 19.45           & 17.43         & 90.24           & 42.27         & 36.95           & 70.49          & 32.53           & 26.54        & 73.58        & 34.44   & 34.45 & 71.74 & 32.17 & 28.84 & 76.51    \\
		FedMA~\cite{wang2020federated}   & 29.68      &  25.85       &  84.68      & 31.17    &   30.78   & 78.81     &  24.82 &  26.75        & 73.05       & 30.95   & 25.15 & 84.28 & 29.15 & 27.13 &  80.20                                                                                                             \\ \hline
		\textbf{FedGPAD}  & 12.73    & 13.36  & 91.25    & 28.69   & 27.55    & 80.58    & 10.97    & 11.11     & 95.34     & 21.95     & 17.91 & 89.85 & \textbf{18.59} & \textbf{17.48} & \textbf{89.25}  \\
	 \bottomrule[1pt]
	\end{tabular}
	\label{tab:STOA_FL}
\end{table*}

In this section, we demonstrate the practicality and generalization ability of the proposed framework in real-world scenario. We first compare the performance of the proposed framework with models trained on data from a single data center. As mentioned above, due to the restriction of data privacy in practice, data cannot be shared among different \textit{data centers}. In this case, \textit{users} will directly obtain a trained model from one of the \textit{data centers}. We report the performance of this baseline in Table~\ref{tab:singleallours} under the label \textbf{Single}. For different choices of user datasets (from O, C, I, M), we report the performance when the model is trained from the remaining datasets independently. 

Rather than obtaining a trained model from a single data center, it is possible for users to obtain multiple trained models from several data centers and fuse their prediction scores during inference, which is also privacy preserving. In this case, we fuse the prediction scores of the trained model from various data centers by calculating the average. The results of this baseline are shown in Table~\ref{tab:singleallours} denoted as \textbf{Fused}. Note that it is impractical for users to carry out feature-level fusion because a classifier is not allowed to be trained based on the fused features through access to any real/spoof data during inference. According to  Table~\ref{tab:singleallours}, fusing scores obtained from different data centers improves the fPAD performance on average. However, this would require higher inference time and computation complexity (of order three for the case considered in this experiment).

On the other hand, Table~\ref{tab:singleallours} illustrates that the average values of all evaluation metrics of the proposed \textbf{FedPAD} framework outperform both baselines. This demonstrates that the proposed FedPAD method is more effective in exploiting fPAD information from multiple data centers. This is because the proposed framework actively combines fPAD information across data centers during training as opposed to the \textbf{fused} baseline. As a result, it is able to generalize better to unseen/novel spoof attacks. Moreover, we further tabulate the performance of the improved \textbf{FedGPAD} framework in Table~\ref{tab:singleallours} and further plot their corresponding ROC curves in four settings corresponding to fPAD datasets in Fig~\ref{fig:rocmain}. The results in Table~\ref{tab:singleallours} and Fig~\ref{fig:rocmain} show that the proposed \textbf{FedGPAD} outperforms the other baselines and significantly improves the performance compared to \textbf{FedPAD}. This demonstrates that the strategy of federated domain disentanglement is extremely effective for the improvement of generalization ability of fPAD under the FL framework and thus a more optimal fPAD model can be obtained based on FedGPAD.

Moreover, we further consider the case where a model is trained with data from all available data centers, which is denoted as \textbf{All} in Table~\ref{tab:singleallours}. Note that this baseline violates the assumption of preserving data privacy, and therefore is not a valid comparison for FedPAD. Nevertheless, it indicates the upper bound of performance for the proposed \textbf{FedPAD} framework. From Table~\ref{tab:singleallours}, it can be seen that the proposed \textbf{FedPAD} framework is only $3.9 \%$ worse than the upper bound in terms of AUC. Specifically, the proposed \textbf{FedGPAD} can even outperform than this baseline \textbf{All} by about $9\%$ improvement in HTER and AUC. This further shows the proposed framework \textbf{FedGPAD} is able to obtain a generalized fPAD model in a privacy persevering manner without sacrificing the fPAD performance. 

\subsection{Comparison with state-of-the-art face presentation attack detection methods}

Although all the current state-of-the-art face presentation attack detection methods exploiting the generalization ability of fPAD do not consider the privacy issue, the comparison with them can still be a useful indicator for the proposed method. Therefore, we further compare the performance of our method with the following state-of-the-art fPAD methods: \textbf{Multi-Scale LBP (MS\_LBP)}~\cite{2011IJCBmstexture} ; \textbf{Binary CNN}~\cite{2014arxivdeepfeat}; \textbf{Image Distortion Analysis (IDA)}~\cite{2015TIFSida}; \textbf{Color Texture}~\cite{2016TIFScolortxt}; \textbf{LBPTOP}~\cite{2014EJIVPlbptop}; \textbf{Auxiliary(Depth Only)}~\cite{2018CVPRauxliary};\textbf{MMD-AAE}~\cite{2018CVPRdgadv}; and \textbf{MADDG}~\cite{Shao_2019_CVPR}; \textbf{DR-MD-Net}~\cite{wang2020cross};\textbf{RFMeta}~\cite{Shao_2020_AAAI}; and \textbf{NAS-Baseline}, \textbf{NAS-Baseline w/ D-Meta}, \textbf{NAS-FAS}, \textbf{NAS-FAS w/ D-Meta}~\cite{yu2020fas}; and \textbf{DC-CDN}~\cite{yu2021dual}. 

We tabulate the results of comparison in Table~\ref{tab:STOA}. From Table~\ref{tab:STOA}, it can be seen that even without getting access to all the source domain data from data centers, the proposed FedGPAD can still reach very close average performance in terms of the metrics of HTER and AUC to the state-of-the-art method NAS-FAS w/ D-Meta~\cite{yu2020fas}, with only about $4 \%$ degradation in HTER and AUC. Specifically, Table~\ref{tab:STOA} shows that the proposed FedGPAD can achieve the best performance in some scenarios, such as the  best HTER results in \textbf{O\&C\&M to I} and the second best HTER performance in \textbf{O\&C\&I to M}. Note that the method NAS-FAS w/ D-Meta~\cite{yu2020fas} not only requires data from all the source domains without privacy preserving, but also leverages very complicated techniques such as neural architecture search (NAS), and cross-domain/type meta-learning, which are all extremely time consuming and difficult to train. In contrast, the proposed FedGPAD framework can be easily implemented and trained. These results further demonstrate that the proposed FedGPAD framework can achieve comparable performance to the state-of-the-art face presentation attack detection methods in a privacy preserving way.

\subsection{Comparison with state-of-the-art federated learning methods}
This section further compares the proposed method with some state-of-the-art FL methods: \textbf{Federated Averaging (FedAvg})~\cite{mcmahan2016communication} and \textbf{Federated Matched Averaging (FedMA})~\cite{wang2020federated}. We tabulate the results of comparison in Table~\ref{tab:STOA_FL}. From experimental results tabulated in Table~\ref{tab:STOA_FL}, it can be seen that FedMA improves the performance compared with vanilla FL method FedAvg for the task of fPAD as it adopts a more advanced network parameters averaging mechanism. Furthermore,  the proposed FedGPAD outperforms current FL methods such as ~\cite{mcmahan2016communication,wang2020federated}. This is because compared with existing FL methods, the federated domain disentanglement strategy proposed in this paper is more able to disentangle the domain-invariant information in the server under the FL framework. This results in a more generalized global fPAD model to deal with various unseen face presentation attacks.

\subsection{Ablation Study}

\begin{table}[htb]
	\renewcommand{\arraystretch}{1}
	\centering	
	\caption{Results of the ablation study corresponding to the loss function.}
	\begin{tabular}{c|c|c|c|c|c}
		\toprule[1pt]
		\multicolumn{3}{c|}{Components} & \multicolumn{3}{c}{\textbf{O\&C\&M to I}} \\ \hline
		$\mathcal{L}\_Diff$   & $\mathcal{L}\_Rec$   & $\mathcal{L}\_Dep$   & HTER(\%)   & EER(\%)   & AUC(\%)  \\ \hline
		& \checkmark        & \checkmark        & 31.71      & 23.56     & 78.94    \\
		\checkmark         &          & \checkmark        & 40.29      & 24.59     & 83.10    \\
		\checkmark         & \checkmark        &          &   27.04         &   27.80       &     73.93     \\ 
		\checkmark         & \checkmark        & \checkmark         &  \textbf{10.97}          &  \textbf{11.11}         &    \textbf{95.34}      \\ 
	\bottomrule[1pt]
	\end{tabular}
	\label{tab:ASloss}
\end{table}

\begin{table*}[htb]
	\renewcommand{\arraystretch}{1}
	\centering	
	\caption{Results of the ablation study corresponding to federated domain disentanglement.}
	\begin{tabular}{c|c|c|c|c|c|c|c|c|c|c|c|c|c|c|c}
		\toprule[1pt]
		\multirow{2}{*}{\textbf{Method}} & \multicolumn{3}{c|}{\textbf{O\&C\&I to M}} & \multicolumn{3}{c|}{\textbf{O\&M\&I to C}} & \multicolumn{3}{c|}{\textbf{O\&C\&M to I}} & \multicolumn{3}{c|}{\textbf{I\&C\&M to O}} & \multicolumn{3}{c}{\textbf{Avg.}}               \\ \cline{2-16} 
		& HTER      & EER      & AUC     & HTER      & EER      & AUC     & HTER      & EER      & AUC     & HTER      & EER      & AUC     & HTER       & EER        & AUC        \\ \hline
		\textbf{FedGPAD w/o Dis}         & 18.55         & 18.14        & 88.87       & 28.11         & 27.29        & 82.06       & 34.14         & 38.56        & 73.31       & 26.41         & 22.44        & 85.08       & 26.80          & 26.61          & 82.33          \\ \hline
		\textbf{FedGPAD}                 & 12.73         & 13.36        & 91.25       & 28.69         & 27.55        & 80.58       & 10.97         & 11.11        & 95.34       & 21.95         & 17.91        & 89.85       & \textbf{18.59} & \textbf{17.48} & \textbf{89.25} \\
	\bottomrule[1pt]
	\end{tabular}
	\label{tab:ASdis}
\end{table*}

We first carry out an ablation study corresponding to the loss in Eq. \ref{equ:loss}, which is composed of four loss components ($\mathcal{L}\_Diff, \mathcal{L}\_Rec, \mathcal{L}\_Dep, \mathcal{L}\_Cls$). $\mathcal{L}\_Cls$ is necessary for classification and thus the ablation study is conducted in terms of the other three losses and the results in the setting of O\&M\&I to C are tabulated in Table \ref{tab:ASloss}. From Table \ref{tab:ASloss}, it can be seen that removing any loss component in the optimization will significantly degrade the performance of the whole framework, which demonstrates that all the added loss components complement each other and all of them contribute to the better overall performance together.

Since federated domain disentanglement is the key strategy proposed in FedGPAD framework, one more ablation study corresponding to this key strategy is conducted in all four settings and the results are shown in Table \ref{tab:ASdis}. Table \ref{tab:ASdis} shows that the average performance of all the three metrics is significantly improved by the strategy of federated domain disentanglement. This further demonstrates the effectiveness of the federated domain disentanglement in the FedGPAD framework.

\subsection{Comparison of different number of data centers}
\begin{table}[htb]
	\renewcommand{\arraystretch}{1}
	\centering	
	\caption{Comparison of different number of data centers}
	\begin{tabular}{c|c|c|c|c|c}
		\toprule[1pt]
		\textbf{Methods}                  & \textbf{Data Centers} & \textbf{User}      & \textbf{HTER} & \textbf{EER} & \textbf{AUC} \\ \hline
		\multirow{3}{*}{\textbf{FedPAD}}  & O\&I                  & \multirow{6}{*}{C} & 49.22              & 49.10             & 51.19             \\
		& O\&M\&I               &                    & 42.27              & 36.95             & 70.49             \\
		& O\&M\&I\&S            &                    & 41.74              & 29.90             & 78.47             \\ \cline{1-2} \cline{4-6} 
		\multirow{3}{*}{\textbf{FedGPAD}} & O\&I                  &                    & 34.17              & 41.78             & 68.74             \\
		& O\&M\&I               &                    & 28.69              & 27.55             & 80.58             \\
		& O\&M\&I\&S            &                    & \textbf{24.20}              & \textbf{23.51}             & \textbf{85.34}             \\ 
		\bottomrule[1pt]
	\end{tabular}
	\label{tab:datacenters1}
\end{table}

\begin{table}[htb]
	\renewcommand{\arraystretch}{1}
	\centering	
	\caption{Comparison of different number of data centers.}
	\begin{tabular}{c|c|c|c|c|c}
		\toprule[1pt]
		\textbf{Methods}                  & \textbf{Data Centers} & \textbf{User}      & \textbf{HTER} & \textbf{EER} & \textbf{AUC} \\ \hline
		\multirow{3}{*}{\textbf{FedPAD}}  & I\&M                  & \multirow{6}{*}{S} & 25.11              & 20.64             & 88.08             \\
		& I\&C\&M               &                    & 29.61              & 14.61             & 93.30             \\
		& I\&C\&M\&O            &                    & 12.45              & \textbf{8.98}     & \textbf{97.18}    \\ \cline{1-2} \cline{4-6} 
		\multirow{3}{*}{\textbf{FedGPAD}} & I\&M                  &                    & 13.67              & 13.85             & 93.72             \\
		& I\&C\&M               &                    & 11.49              & 11.59             & 95.54             \\
		& I\&C\&M\&O            &                    & \textbf{9.72}      & 9.86              & 96.37             \\ 
	\bottomrule[1pt]
	\end{tabular}
	\label{tab:datacenters2}
\end{table}

\begin{table*}
	\renewcommand{\arraystretch}{1}
	\centering	
	\caption{Effect of using different types of spoof attacks.}
	\begin{tabular}{c|c|c|c|c|c}
		\toprule[1pt]
		\textbf{Methods}                 & \textbf{Data Centers}  & \textbf{User}    & \textbf{HTER (\%)} & \textbf{EER (\%)} & \textbf{AUC (\%)} \\ \hline
		\multirow{2}{*}{\textbf{Single}} & I (Print)              & M (Print, Video) & 38.82              & 33.63             & 72.46             \\ \cline{2-6} 
		& O (Video)              & M (Print, Video) & 35.76              & 28.55             & 78.86             \\ \hline
		\textbf{Fused}                     & I (Print) \& O (video) & M (Print, Video) & 35.22              & 25.56            & 81.54             \\ \hline		
		\textbf{FedPAD}                    & I (Print) \& O (video) & M (Print, Video) & 30.51    & 26.10    & \textbf{84.82}    \\ \hline
		\textbf{FedGPAD}                    & I (Print) \& O (video) & M (Print, Video) & \textbf{28.19}     & \textbf{23.01}    & 84.75   \\ \bottomrule[1pt]
	\end{tabular}
	\label{tab:2Dtype}
\end{table*}

\begin{figure}[!htb]
	\begin{center}
		\includegraphics[ width=0.5\linewidth]{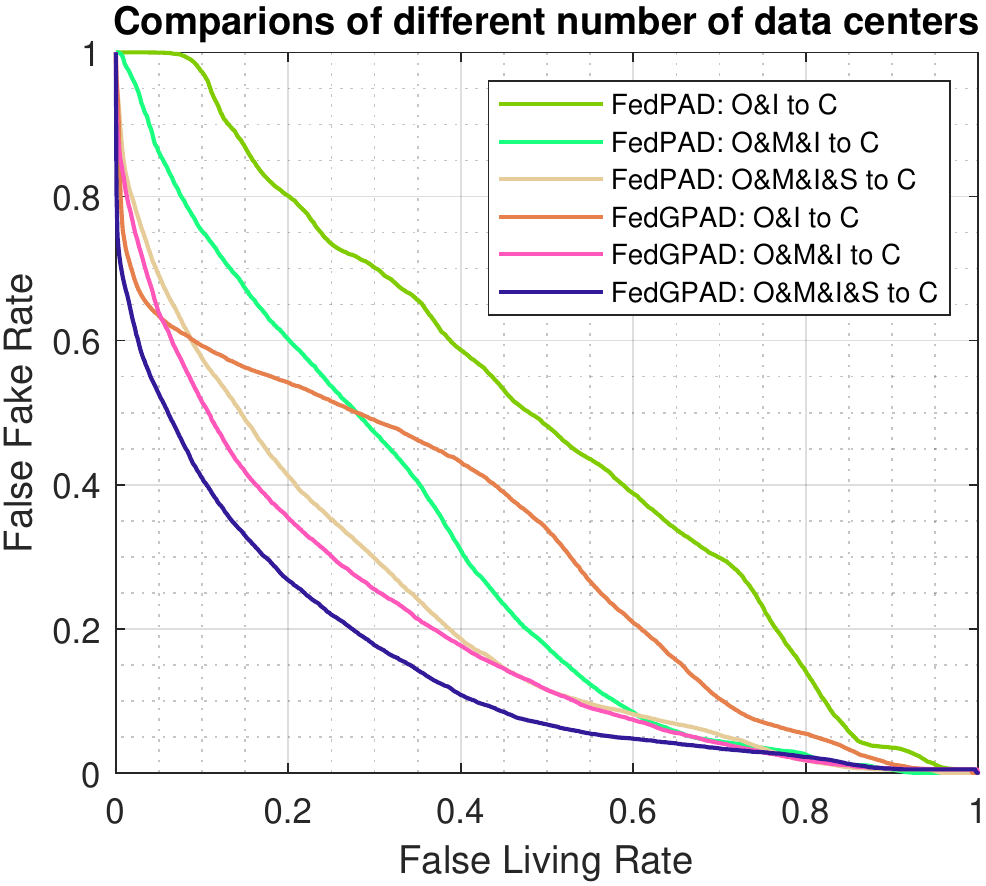}
	\end{center}
	\caption{ROC curves of different number of data centers.}
	\label{fig:numberofdatacenter}
\end{figure}

\begin{figure*}[!htb]
	\begin{center}
		\includegraphics[ width=0.9\linewidth]{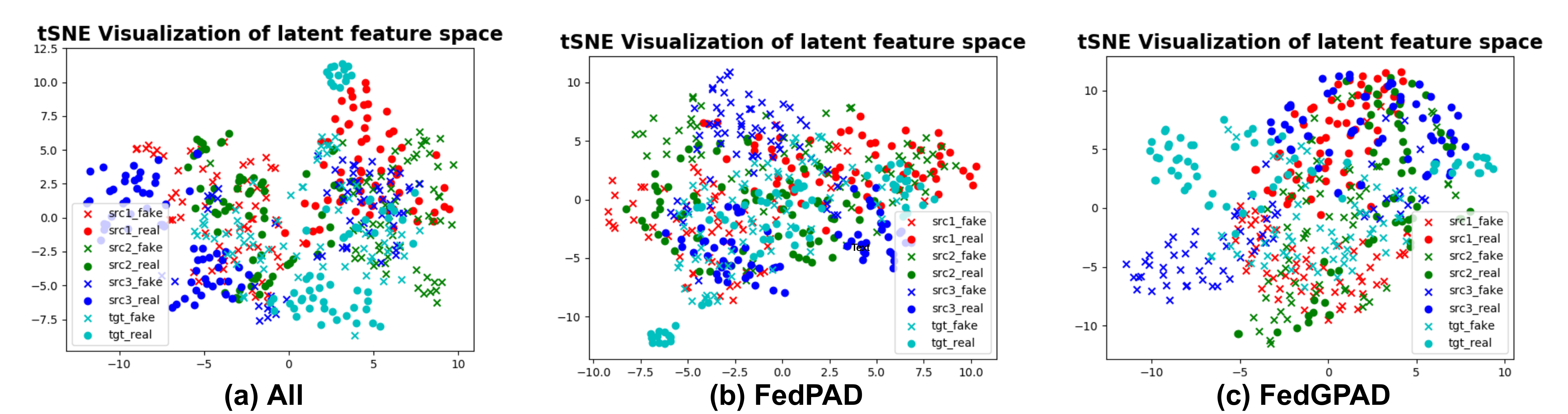}
	\end{center}
	\caption{tSNE visualization of features in the setting of O\&C\&M to I, where data centers are O, C and M, and data from dataset I is presented to the user. For baseline All and the proposed FedPAD framework, src1\_real/fake, src2\_real/fake, src3\_real/fake means the features extracted by fPAD model in the server when the training data of real and spoof faces are from data centers O, C and M respectively, while tgt\_real/fake denotes the features extracted by fPAD model in the server when the testing data of real and spoof faces are from dataset I. For the proposed FedGPAD framework, the features are extracted by the domain-invariant feature extractor in the server.}
	\label{fig:tsne_OCM2I}
\end{figure*}

\begin{figure*}[!htb]
	\begin{center}
		\includegraphics[ width=0.9\linewidth]{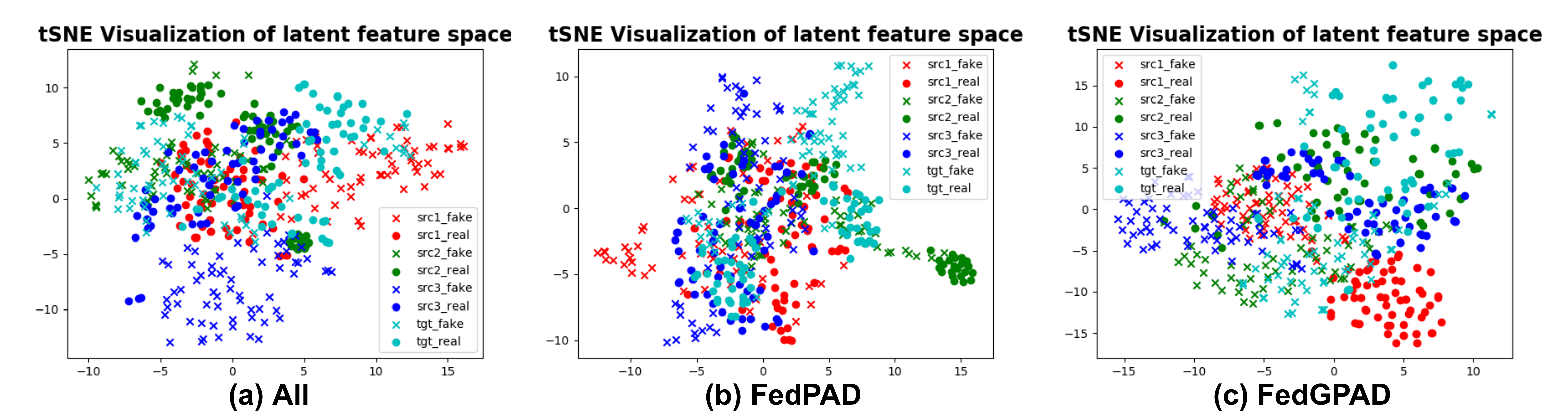}
	\end{center}
	\caption{tSNE visualization of domain-invariant features in the setting of O\&C\&I to M.}
	\label{fig:tsne_OCI2M}
\end{figure*}

In this section, we investigate the importance of having more  data centers during training for both FedPAD and FedGPAD frameworks. Different data centers exploit different characteristics and scenarios of face presentation attacks. Therefore, we expect aggregating fPAD models from more data centers will enable the aggregated model to summarize more knowledge from distinct characteristics and scenarios of face presentation attacks, and thus learn domain-invariant features with better generalization to unseen attacks. To verify this point, we increase the number of data centers in the proposed FedPAD and FedGPAD frameworks and report the results in Tables~\ref{tab:datacenters1} ~\ref{tab:datacenters2} and Fig.~\ref{fig:numberofdatacenter}. The experiments are carried out using five datasets (O, M, I, C, S). In Fig.~\ref{fig:numberofdatacenter} and Table~\ref{tab:datacenters1}, we select dataset C as the data presented to the user and the remaining datasets as the data centers for training the fPAD model with our frameworks. We increase the number of data centers from 2 to 4. As shown in Fig.~\ref{fig:numberofdatacenter} and Table~\ref{tab:datacenters2}, another experiment is carried out with a different combination of the same five datasets by altering the user as S. From the curve in Fig.~\ref{fig:numberofdatacenter} and Tables~\ref{tab:datacenters1} ~\ref{tab:datacenters2}, it can be seen that evaluation results improve when the number of data centers increases in both FedPAD and FedGPAD frameworks. This demonstrates that increasing the number of data centers in the proposed FedPAD framework can improve the performance. Moreover, FedGPAD can achieve better performance when the number of data centers go from 2 to 4, which can further demonstrate the superiority of the proposed FedGPAD framework.

\subsection{Generalization ability to various 2D spoof attacks}

In reality, due to limited resources, one data center may only have limited types of 2D attacks. However, various 2D attacks may appear to the users. Assume that one data center collects one particular type of 2D attack such as print attack or video-replay attack. As illustrated in Table~\ref{tab:2Dtype}, first, we select real faces and print attacks from dataset I and real faces and video-replay attacks from dataset O to train a fPAD model, respectively and evaluate them on dataset M (containing both print attacks and video-replay attacks). In both considered cases as shown in Table~\ref{tab:2Dtype}, the corresponding trained models cannot generalize well to dataset M which contains the additional types of 2D attacks compared to dataset I and O, respectively. This tendency can be alleviated when the prediction scores of two independently trained models on both types of attacks are fused as shown in Table~\ref{tab:2Dtype}. Comparatively, FedPAD method obtains a performance gain of $4.71\%$ in HTER and $3.3\%$ in AUC compared to score fusion. Furthermore, FedGPAD improves the EER and HTER performances and achieves comparable AUC results compared to FedPAD. This experiment demonstrates that carrying out FedPAD and FedGPAD frameworks among data centers with different types of 2D spoof attacks can improve the generalization ability of the trained fPAD model to various 2D spoof attacks.

\subsection{Visualization of latent features}

To better understand the proposed algorithm, in Figures~\ref{fig:tsne_OCM2I} and~\ref{fig:tsne_OCI2M}, we visualize the features extracted from the fPAD models in baseline All and FedPAD, and from the domain-invariant feature extractor in FedGPAD framework respectively using tSNE visualization~\cite{maaten2008visualizing}. From Fig~\ref{fig:tsne_OCM2I} and~\ref{fig:tsne_OCI2M}, it can be seen that in baseline All and the proposed FedPAD framework based on the vanilla FL, features between real and spoofing faces are not well separated, and feature distributions between data centers and users are not aligned. In contrast, in the proposed FedGPAD framework, the generated domain-invariant features can be well separated between real and spoof faces from both data centers and users and a better class boundary between real and spoof faces can be achieved by FedGPAD for data from both data centers and the user. Moreover, we can clearly observe that the feature distributions between data centers and users can be well aligned. This means that the domain-invariant feature extractor in the proposed FedGPAD framework can truly extract  invariant features between data centers during training and users during testing. This means these features can generalize well to the unseen attacks presented to the users.

\section{Conclusion}
In this paper, we presented FedPAD, a FL-based framework targeting application of fPAD with the objective of obtaining generalized fPAD models while preserving data privacy. Through communications between \textit{data centers} and the \textit{server}, a global fPAD model is obtained by iteratively aggregating the model updates from various \textit{data centers}. Local private data is not accessed during this  process. To equip the fPAD model with better generalization ability to unseen attacks from users, we further propose a Federated Generalized Face Presentation Attack Detection (FedGPAD) framework. A federated domain disentanglement strategy is exploited in this framework which disentangles the domain-invariant part of fPAD models among the local data centers and uses them to communicate with server for model aggregation. As such, a global fPAD model with better domain-invariant information will be aggregated in the server, which is more able to generalize well the unseen attacks in the users. Extensive experiments are carried out to demonstrate the effectiveness of the proposed framework which provide various insights regarding FL for fPAD.

%\vfill

%\enlargethispage{-2in}
{
\Large
\bibliographystyle{ieee}
\bibliography{ref}
}

\end{document}